\documentclass[journal]{IEEEtran}
\usepackage{amsmath,amsfonts,amssymb}
\usepackage{newtxtext,newtxmath}
\usepackage{algorithmic}
\usepackage{algorithm}
\usepackage{array}
\usepackage{booktabs}
\usepackage{tabularx}
\usepackage{adjustbox}
\usepackage{caption}
\DeclareCaptionLabelFormat{dapgfiglabel}{Fig.#2}
\usepackage{subfig}
\usepackage{textcomp}
\usepackage{stfloats}
\usepackage{url}
\usepackage{verbatim}
\usepackage{graphicx}
\usepackage{lettrine}
\graphicspath{{./}}
\usepackage{cite}
\usepackage{bm}
\newcommand{\tnrinline}[1]{$\mathrm{#1}$}
\renewcommand{\IEEEPARstart}[2]{%
\lettrine[lines=2,lhang=0,nindent=0em,slope=0em,findent=0.15em]{\bfseries #1}{\MakeUppercase{#2}}%
}
\makeatletter
\def\section{\@startsection{section}{1}{\z@}{3.0ex plus 1.5ex minus 1.5ex}%
{0.7ex plus 1ex minus 0ex}{\normalfont\normalsize\rmfamily\centering\scshape}}
\def\subsection{\@startsection{subsection}{2}{\z@}{3.5ex plus 1.5ex minus 1.5ex}%
{0.7ex plus .5ex minus 0ex}{\normalfont\normalsize\rmfamily\itshape}}
\makeatother
\newcolumntype{Y}{>{\centering\arraybackslash}X}
\newcommand{\dapgtablefontsize}{\small}
\newcommand{\dapgtablewidth}{\textwidth}
\newcommand{\dapgtableformat}{\dapgtablefontsize\rmfamily}

\begin{document}

\title{DAPGNet: Dynamic Adaptive Physics-Guided Graph Diffusion Network for Hyperspectral Image Classification}

\author{Pengkun~Wang,
        Weijia~Cao,
        Ning~Wang,
        and Xiaofei~Yang%
\thanks{Pengkun Wang and Weijia Cao contributed equally to this work.}
\thanks{(Corresponding author: Weijia Cao, Ning Wang, and Xiaofei Yang.)}
\thanks{Pengkun Wang is with the National Engineering Laboratory for Satellite Remote Sensing Applications, Aerospace Information Research Institute, Chinese Academy of Sciences, Beijing 100094, China, and also with the University of Chinese Academy of Sciences, Beijing 100049, China (e-mail: wangpengkun25@mails.ucas.ac.cn).}
\thanks{Weijia Cao is with the Aerospace Information Research Institute, Chinese Academy of Sciences, Beijing 100094, China (e-mail: caowj@aircas.ac.cn).}
\thanks{Ning Wang is with the National Engineering Laboratory for Satellite Remote Sensing Applications, Aerospace Information Research Institute, Chinese Academy of Sciences, Beijing 100094, China (e-mail: wangning@aircas.ac.cn).}
\thanks{Xiaofei Yang is with Guangzhou University, Guangzhou 510006, China (e-mail: xiaofeiyang@gzhu.edu.cn).}
}

\markboth{}{}

\IEEEaftertitletext{\vspace{-0.9\baselineskip}}
\maketitle

\begin{abstract}
Hyperspectral image (HSI) classification requires reliable pixel-relation modeling under spectral variability, mixed pixels, and heterogeneous boundaries. Existing graph-based HSI classifiers usually construct graph topology from spatial proximity, superpixel connectivity, or learned feature affinity. However, the spectral physical prior carried by contiguous bands has limited influence on topology estimation and message propagation. This paper presents DAPGNet, a dynamic adaptive physics-guided graph diffusion network that injects a structure-constrained physical prior into relation-level graph learning. DAPGNet first encodes contiguous spectral responses into node-wise multiscale physical-prior representations. A two-stage graph constructor then combines spectral--spatial affinity, physical-prior consistency, and spatial distance to form a physical-prior-aware sparse topology. During graph diffusion, learned edge weights are transformed into additive attention biases, while a physical gate performs node-wise and feature-wise interpolation between graph-aggregated features and projected physical-prior features. Cross-scale fusion integrates node states from different diffusion depths, and the network is optimized with main classification, auxiliary supervision, and second-order spectral smoothness regularization. Experiments on Indian Pines, WHU-Hi-LongKou, Houston2013, and Houston2018 show that DAPGNet achieves the best OA, AA, and Kappa among representative CNN-, Transformer-, Mamba-, and graph-based baselines. It improves AA over the strongest competing method by 3.64--7.31 percentage points across the four datasets. Ablation and sensitivity analyses further support the complementary effects of physical-prior extraction, prior-aware topology construction, physics-gated propagation, and spectral smoothness regularization.
\end{abstract}


\section{Introduction}
\IEEEPARstart{H}{yperspectral} imaging records a scene over hundreds of contiguous narrow bands, providing detailed spectral measurements associated with material composition and surface properties \cite{hongHyperspectralImaging2026,ahmadComprehensiveSurveyHyperspectral2025}. The ordered bands also exhibit structural regularities that support material discrimination beyond visual appearance. In this article, the term spectral physical prior refers to regularities inherited from contiguous hyperspectral measurements, including adjacent-band continuity, local spectral smoothness, and multiscale spectral variation. Accordingly, the physical prior in this work is confined to structure-constrained regularities derived from ordered contiguous spectra. Hereafter, physical prior is used for brevity. These regularities provide complementary cues when visually similar objects exhibit distinct spectral responses. Preserving this prior is especially valuable under limited supervision, where noise, spectral variability, and heterogeneous boundaries induce unstable correlations in purely data-driven representations.

HSI classifiers have progressively expanded their spectral--spatial modeling capacity. Convolutional architectures capture local patterns with one-, two-, or three-dimensional kernels \cite{benHamida3DDeepLearning2018,royHybridSNExploring2020}; Transformers model nonlocal interactions through spectral or spectral--spatial tokens \cite{hongSpectralFormerRethinking2022}; and Mamba-based architectures provide efficient sequence modeling over long spectral and spatial paths \cite{liMambaHSISpatialSpectral2024}. Information exchange in these architectures follows regular receptive fields, token sequences, or scanning paths. Mixed spectra and irregular boundaries additionally call for sparse, sample-adaptive pixel relations, especially when illumination variation, noise, or transitional regions produce misleading similarities.

Graph neural networks (GNNs) address this relational requirement by organizing pixels or regions as nodes and propagating information along selected edges \cite{hongGraphConvolutionalNetworks2020,zhaoReviewHyperspectralImage2025}. Graph convolution, graph attention, and dynamic graph learning have improved HSI classification through superpixel relations, feature-dependent aggregation, and adaptive topology refinement \cite{liuCNNEnhancedGraph2021,dongWeightedFeatureFusion2022,wanMultiscaleDynamicGraph2020}. In this formulation, topology specifies which nodes exchange information, whereas aggregation weights determine the relative contribution of their messages. Both stages therefore affect graph reliability. Existing HSI graphs are commonly derived from spatial adjacency, superpixel connectivity, learned feature affinity, or attention scores. Around mixed pixels and heterogeneous boundaries, these criteria connect nodes with similar observations but different material compositions, allowing irrelevant neighborhood messages to contaminate node representations during message propagation.

Physics-guided learning provides a principled route for incorporating domain knowledge into neural models. Equation-driven approaches impose governing equations or residual constraints \cite{raissiPhysicsInformedNeural2019,karniadakisPhysicsInformedMachine2021}, whereas HSI classification rarely offers a universal scene-level equation. Structure-constrained approaches instead incorporate spectral ordering, subpixel mixing, or measurement characteristics into model design. Existing HSI studies have mainly introduced such information into spectral representation, abundance estimation, reconstruction, attention masks, or state evolution \cite{hongEndmemberGuidedUnmixing2022,zhouPhysicsinformedInteractiveNetwork2024,hanSubpixelSpectralVariability2025,zhangPIMambaPhysicsInformedMamba2026}. In graph-based HSI classification, physical priors seldom participate directly in topology construction or regulate message propagation.

These observations reveal three related limitations. First, the structural manifestations of the physical prior span different local band ranges, calling for complementary spectral receptive fields and spectral attention when forming node-wise prior representations. Second, even when a physical prior is encoded as a node representation, topology is still predominantly determined by spatial proximity or learned feature affinity, leaving the prior weakly involved in deciding which nodes should be connected. Third, once unreliable edges are introduced, conventional graph layers mainly depend on data-driven aggregation weights, allowing irrelevant neighborhood information to accumulate across multiple propagation steps. Addressing these limitations requires a unified framework that derives multiscale physical priors, incorporates them into topology construction, and maintains their guidance throughout subsequent graph learning.

To this end, we propose DAPGNet, a dynamic adaptive physics-guided graph diffusion network for HSI classification. A multiscale spectral physical prior encoder first characterizes contiguous spectral patterns over complementary receptive ranges and produces node-wise physical-prior representations. A two-stage graph constructor then combines spectral--spatial affinity, physical-prior consistency, and spatial distance to screen candidate neighbors and estimate sparse edge weights. During graph diffusion, the learned edge weights are converted into additive attention biases, while a physical gate adaptively balances graph-aggregated information and projected physical-prior features at each layer. The physical prior consequently remains active in node representation, topology construction, attention regulation, and layer-wise graph diffusion.

The main contributions of this article are summarized as follows.
\vspace{-0.35\baselineskip}
\begin{itemize}
    \setlength{\itemsep}{0pt}
    \setlength{\parsep}{0pt}
    \setlength{\parskip}{0pt}
    \item We develop a multiscale spectral physical prior encoder that characterizes contiguous spectral patterns over complementary receptive ranges. Spectral attention integrates these patterns into node-wise physical-prior representations shared by subsequent graph construction and graph diffusion.
    \item We propose a physical-prior-aware adaptive graph construction mechanism that jointly considers spectral--spatial representations, physical-prior consistency, and spatial relations. This design introduces the physical prior directly into topology decisions, providing an additional relation cue around heterogeneous boundaries and transitional regions.
    \item We introduce physics-gated graph diffusion with edge-biased sparse attention. Learned topology weights regulate neighborhood attention, and a node-wise physical gate integrates graph-aggregated information with projected physical-prior features during each diffusion layer, maintaining physical-prior guidance beyond graph construction.
    \item Extensive experiments on four benchmark datasets validate the overall effectiveness of DAPGNet, quantify the complementary contributions of its components, characterize the effects of training ratio and patch size, and assess computational practicality.
\end{itemize}

\section{Related Work}
\subsection{Graph-Based Relational Modeling for HSI Classification}
Graph-based HSI classifiers model irregular relations among pixels, superpixels, or regions. miniGCN reduces the burden of full-graph training through minibatch graph convolution while retaining graph relations defined before feature propagation \cite{hongGraphConvolutionalNetworks2020}. CEGCN combines convolutional and graph branches to integrate pixel-level and superpixel-level representations, with their complementary information fused at the feature level \cite{liuCNNEnhancedGraph2021}. WFCG further fuses convolutional representations with superpixel-based graph-attention features and uses learned attention coefficients to aggregate neighboring nodes \cite{dongWeightedFeatureFusion2022}. This progression extends HSI classifiers from fixed graph convolution to feature-level branch fusion and attention-based neighborhood weighting.

Later studies increase graph flexibility through dynamic, multiview, multiscale, and global relation modeling. FDGC updates graph relations within a parallel graph--convolution architecture \cite{liuFastDynamicGraph2022}. ACSS-GCN constructs separate spatial and spectral graphs and exchanges their information through adaptive cross-attention \cite{yangAdaptiveCrossAttentionDriven2022}. MDGCN constructs and refines relations at multiple scales to aggregate complementary neighborhood information \cite{wanMultiscaleDynamicGraph2020}, while SSGRN enlarges the relational context through global spectral--spatial graph reasoning \cite{wangSpectralSpatialGlobal2023}. These designs improve the spatial extent and adaptivity of graph reasoning, although their relation estimates remain learned primarily from observations and intermediate features.

\subsection{Structure-Aware Global Dependency Modeling With Graphs and Sequence Models}
Graph structures have also been integrated with Transformer and state-space architectures to couple relational reasoning with long-range dependency modeling. GraphGST introduces graph-generated structural information and positional encoding into Transformer token interactions, linking graph context with self-attention-based global modeling \cite{suGraphGSTGraphGenerative2024}. GraphMamba combines graph structure learning with vision Mamba so that learned spatial relations and long-range sequence processing contribute to a shared representation \cite{yangGraphMambaEfficient2024}. Their graph components serve primarily as structural context for global feature interaction.

Hybrid architectures further combine complementary graph and sequence branches. DBMGNet employs a dual-branch Mamba--GCN design for long-range sequential dependencies and non-Euclidean graph relations, followed by cross-branch feature integration \cite{wangDBMGNetDualbranchMambaGCN2025}. GAST couples a graph-augmented spectral--spatial Transformer with adaptive gated fusion \cite{keskinGASTGraphaugmentedSpectralspatial2026}. Beyond explicit graph hybrids, Hi-RWKV introduces structure-guided recurrent modeling through boundary-aware spatial modulation and spectral identity-driven channel mixing for linear-complexity context aggregation \cite{wangHiRWKVHierarchical2026}.

Together, these architectures strengthen structural awareness, long-range dependency modeling, and computational efficiency, while their interaction and gating signals remain primarily data-driven.

\subsection{Physics-Informed and Prior-Guided Hyperspectral Learning}
For the purpose of this review, prior-guided hyperspectral methods are organized by where domain information enters the learning process. Equation-driven learning embeds governing equations or their residuals into optimization when an explicit physical model is available \cite{raissiPhysicsInformedNeural2019,karniadakisPhysicsInformedMachine2021}. When a universal scene-level equation is unavailable, structure-constrained designs encode spectral ordering, subpixel composition, or measurement characteristics into the model architecture. At the representation level, SpectralFormer preserves adjacent-band organization through group-wise spectral embedding \cite{hongSpectralFormerRethinking2022}, while PhISM represents hyperspectral observations with continuous basis functions and learns compact, interpretable spectral components \cite{gawrysiakPhysicsInformedSpectral2025}. Across these designs, spectral organization is introduced mainly at the representation stage.

Model-driven approaches introduce physically meaningful variables into feature learning and reconstruction. The complementarity between discriminative classification and spectral unmixing connects class prediction with subpixel interpretation \cite{liComplementarityDiscriminative2015}. EGU-Net incorporates endmembers and abundance constraints into self-supervised hyperspectral unmixing \cite{hongEndmemberGuidedUnmixing2022}, and S2VNet integrates endmember abundance, spectral variability, and nonlinear mixture characteristics into HSI classification \cite{hanSubpixelSpectralVariability2025}. In these formulations, endmembers, abundances, and variability parameters provide structured intermediate representations. Their main roles lie in spectral reconstruction, feature enhancement, abundance estimation, or auxiliary supervision.

Physical information has also been embedded more deeply into attention and state evolution. PI$^{2}$Net derives component-aware attention masks from abundance features and uses class-wise endmember activation to couple its two streams \cite{zhouPhysicsinformedInteractiveNetwork2024}. PIMamba constrains state evolution through diffusion-inspired spectral--spatial dynamics and associated physical losses \cite{zhangPIMambaPhysicsInformedMamba2026}. These designs allow physical information to regulate feature interaction or state transition.

Graph-based methods improve relational learning but mainly rely on data-driven topology and propagation. Prior-guided methods incorporate physical information primarily at representation, reconstruction, physical-variable, attention, or state-transition levels. The integration of physical priors with both graph topology construction and message propagation remains limited. DAPGNet addresses this gap through physical-prior-aware topology estimation, edge-biased attention, and physics-gated graph diffusion.

\section{Proposed Methodology}

\subsection{Overview of DAPGNet}
For each target pixel, a local HSI patch $\mathbf{X}\in\mathbb{R}^{S\times S\times C}$ is extracted, where $S$ is the spatial side length and $C$ is the number of spectral bands. The $N=S^2$ spatial positions within the patch are treated as graph nodes, and $\boldsymbol{x}_i\in\mathbb{R}^{C}$ denotes the spectral vector of node $i$. The class label of the center pixel is used to supervise patch-level classification. Throughout this section, bold uppercase letters denote matrices or tensors, bold lowercase letters denote vectors, italic letters denote scalars or indices, and calligraphic letters denote functions. The batch dimension is omitted for clarity.

\begin{figure*}[!t]
    \centering
    \includegraphics[width=\textwidth]{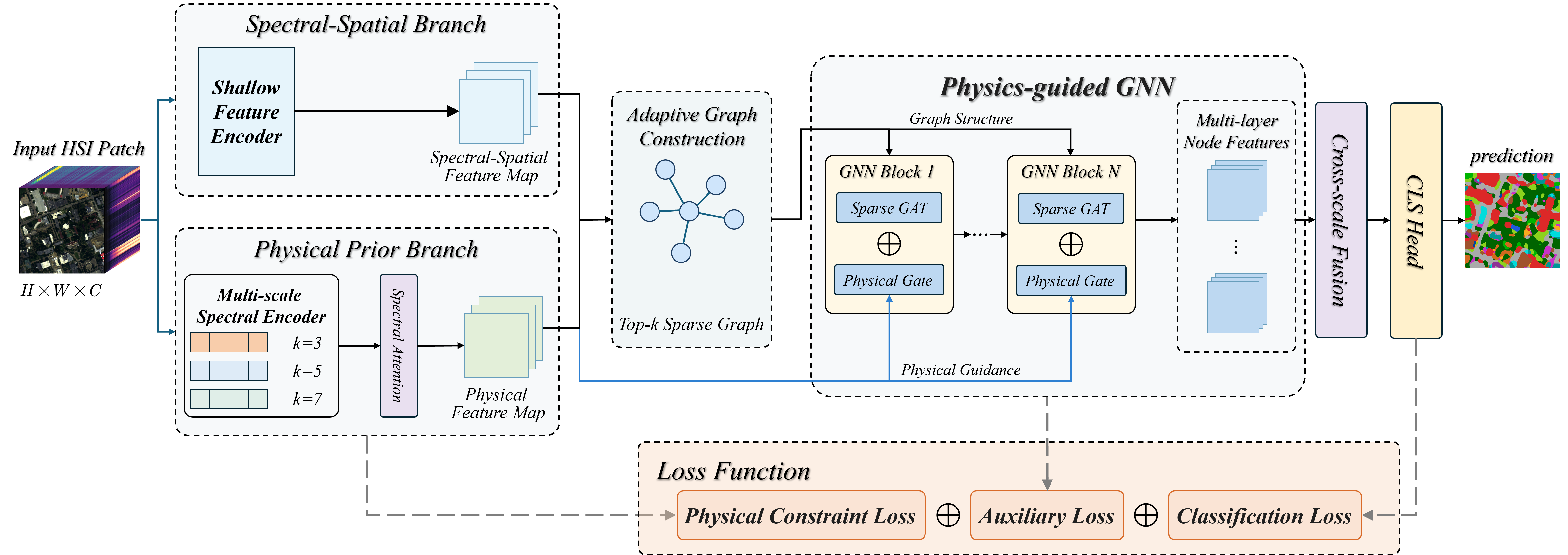}
    \caption{Overall architecture of DAPGNet.}
    \label{fig:framework}
\end{figure*}

As illustrated in Fig.~\ref{fig:framework}, DAPGNet employs two parallel branches to derive complementary node representations from $\mathbf{X}$. A lightweight shallow feature encoder, composed of standard convolution and depthwise separable convolution, extracts local spectral--spatial context while preserving the $S\times S$ spatial layout. Its feature map is reshaped into $\mathbf{F}\in\mathbb{R}^{N\times M}$, where $M$ is the node-feature dimension and the $i$th row $\boldsymbol{f}_i\in\mathbb{R}^{M}$ is used for adaptive graph construction and graph-state initialization. In parallel, the multiscale spectral physical prior encoder processes each contiguous node spectrum over complementary spectral ranges and produces $\mathbf{P}\in\mathbb{R}^{N\times M_p}$, where $M_p$ is the physical-prior dimension and the $i$th row $\boldsymbol{p}_i\in\mathbb{R}^{M_p}$ represents the physical prior of the same node.

The two representations provide complementary information: $\mathbf{F}$ summarizes locally learned spectral--spatial content, and $\mathbf{P}$ encodes structure-constrained regularities along the contiguous spectral dimension. Together with the normalized coordinate matrix $\mathbf{R}\in\mathbb{R}^{N\times2}$, they define the physical-prior-aware graph matrix
\begin{equation}
\mathbf{A}_{\mathrm{phys}}
=
\mathcal{G}(\mathbf{F},\mathbf{P},\mathbf{R}),
\label{eq:overview-graph}
\end{equation}
where $\mathcal{G}$ denotes the adaptive graph construction function. The nonzero entries of $\mathbf{A}_{\mathrm{phys}}$ encode sparse node connectivity and edge strength for subsequent attention regulation.

The shallow node representation $\mathbf{F}$ is linearly mapped to the initial graph state $\mathbf{H}^{(0)}\in\mathbb{R}^{N\times M}$. DAPGNet then performs $L$ successive physics-gated graph diffusion layers, where $L$ is the total diffusion depth, and obtains $\mathbf{H}^{(1)},\ldots,\mathbf{H}^{(L)}$. Within each layer, the edge weights in $\mathbf{A}_{\mathrm{phys}}$ provide additive attention biases for neighborhood diffusion, while $\mathbf{P}$ serves as a node-wise reference for the physical gate. Finally, $\mathbf{H}^{(0)},\ldots,\mathbf{H}^{(L)}$ are fused across diffusion depths and aggregated to predict the class of the center pixel. The physical prior thereby remains active in graph topology construction, attention regulation, and layer-wise graph diffusion.
\subsection{Multiscale Spectral Physical Prior Encoder}
The input associated with node $i$ is the complete ordered spectrum $\boldsymbol{x}_i\in\mathbb{R}^{C}$, which contains all $C$ bands at that spatial position and follows the band sequence of the hyperspectral sensor. This contiguous sequence exhibits inter-band continuity, local spectral smoothness, and spectral variations distributed over different local ranges \cite{bioucasdiasHyperspectralUnmixing2012}. The encoder transforms these regularities into a node-wise physical-prior representation that subsequently guides graph topology construction and graph diffusion.

\begin{figure}[!t]
    \centering
    \includegraphics[width=\columnwidth]{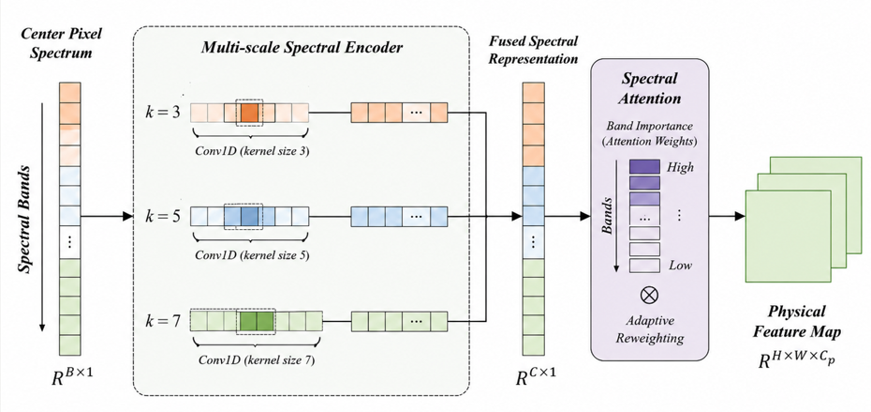}
    \caption{Architecture of the multiscale spectral physical prior encoder.}
    \label{fig:multiscale_encoder}
\end{figure}

Discriminative spectral variations span different local band ranges across land-cover categories, and a single receptive range provides incomplete coverage of these heterogeneous patterns. As illustrated in Fig.~\ref{fig:multiscale_encoder}, $\boldsymbol{x}_i$ is processed in parallel by three one-dimensional spectral branches whose first convolutional kernels have widths 3, 5, and 7. These branches characterize fine adjacent-band variations, intermediate local continuity, and relatively broad local spectral patterns, respectively.

Each branch produces a response sequence distributed along the spectral dimension, retaining band order while indicating the activation of local patterns at different positions. Global average pooling along the band dimension converts each response sequence into a fixed-length branch descriptor, keeping the subsequent fusion form independent of $C$ and maintaining a fixed fusion dimensionality for datasets with different band counts. The three branch descriptors are then concatenated to preserve complementary information from all receptive ranges before adaptive fusion.

A component-wise spectral attention module reweights the concatenated multiscale descriptor, adaptively combining contributions from all three branches for each node. The weighted descriptor is mapped to an $M_p$-dimensional physical-prior space to obtain $\boldsymbol{p}_i\in\mathbb{R}^{M_p}$. In parallel, the unweighted descriptor follows a direct projection path and is added to the attention-guided output, providing a residual path for retaining the original multiscale descriptor.

Stacking the physical-prior features of all nodes gives
\begin{equation}
\mathbf{P}
=
[\boldsymbol{p}_1,\boldsymbol{p}_2,\ldots,\boldsymbol{p}_N]^{\top}
\in\mathbb{R}^{N\times M_p}.
\label{eq:prior-matrix}
\end{equation}
During training, a second-order spectral smoothness term regularizes the pre-pooling branch responses, encouraging locally continuous physical-prior features along the ordered band dimension. The resulting $\mathbf{P}$ participates in candidate-node selection and edge-weight estimation, and its layer-specific projection enters the physical gate at every graph diffusion layer.

\subsection{Physical-Prior-Aware Adaptive Graph Construction}
Given the spectral--spatial representation $\mathbf{F}$ and the physical-prior representation $\mathbf{P}$, this module constructs an input-adaptive graph for each HSI patch. Fixed spatial neighborhoods provide limited flexibility in heterogeneous regions, where adjacent pixels exhibit distinct spectral--spatial content and physical-prior responses. The connectivity of each node pair is therefore evaluated from spectral--spatial consistency, physical-prior consistency, and spatial proximity, allowing the graph topology to follow the local composition of the input patch.

Let $\boldsymbol{r}_i\in\mathbb{R}^{2}$ denote the spatial coordinate of node $i$, corresponding to row $i$ of $\mathbf{R}$. The normalized spatial distance between nodes $i$ and $j$ is defined as
\begin{equation}
\delta_{ij}
=
\frac{\lVert\boldsymbol{r}_i-\boldsymbol{r}_j\rVert_2}{\delta_{\max}},
\label{eq:normalized-spatial-distance}
\end{equation}
where $\delta_{\max}$ is the maximum Euclidean distance between any two nodes within the current patch, yielding $\delta_{ij}\in[0,1]$. Before coarse screening, residual multihead self-attention contextualizes $\mathbf{F}$, while lightweight projection layers map $\mathbf{F}$ and $\mathbf{P}$ into compact screening spaces. The operator $\operatorname{sim}(\cdot,\cdot)$ in the following equation denotes cosine similarity after the corresponding projection and $\ell_2$ normalization. A coarse relation score is then computed by
\begin{equation}
s_{ij}
=
\eta_f\operatorname{sim}(\boldsymbol{f}_i,\boldsymbol{f}_j)
+\eta_p\operatorname{sim}(\boldsymbol{p}_i,\boldsymbol{p}_j)
+\eta_{\delta}\delta_{ij},
\label{eq:coarse-relation-score}
\end{equation}
The coefficients $\eta_f$, $\eta_p$, and $\eta_{\delta}$ control the contributions of spectral--spatial similarity, physical-prior similarity, and spatial distance, respectively. A larger $s_{ij}$ indicates greater suitability for candidate connection.

For each node $i$, self-connection is excluded during coarse screening, and the indices of the $m$ largest scores form the candidate set
\begin{equation}
\mathcal{C}_i
=
\operatorname{TopK}\!\left(\{s_{ij}\mid j\ne i\},m\right),
\label{eq:candidate-neighborhood}
\end{equation}
where $\operatorname{TopK}(\cdot,m)$ returns the indices of the $m$ largest elements, $m>k$, and $k$ is the final neighborhood size. A learnable edge scorer is subsequently applied to the retained candidate pairs:
\begin{equation}
e_{ij}
=
\sigma\!\left(
\operatorname{MLP}\!\left(
\boldsymbol{f}_i,\boldsymbol{f}_j,
\boldsymbol{p}_i,\boldsymbol{p}_j,
\delta_{ij}
\right)
\right)
\label{eq:candidate-edge-weight}
\end{equation}
where the listed inputs are concatenated before entering the MLP, and $\sigma$ is the sigmoid function. The resulting $e_{ij}\in(0,1)$ quantifies the directed connection strength from node $i$ to node $j$. The physical-prior representation consequently contributes to both candidate screening and learnable edge-weight estimation.

The $k$ candidate nodes with the largest learned edge weights define the final neighborhood
\begin{equation}
\mathcal{N}_i
=
\operatorname{TopK}\!\left(\{e_{ij}\mid j\in\mathcal{C}_i\},k\right).
\label{eq:final-neighborhood}
\end{equation}
The selected edges are organized into the sparse edge-weight matrix
\begin{equation}
E_{ij}
=
\begin{cases}
e_{ij}, & j\in\mathcal{N}_i,\\[1mm]
0, & \text{otherwise},
\end{cases}
\qquad
\mathbf{E}\in\mathbb{R}^{N\times N}.
\label{eq:sparse-edge-matrix}
\end{equation}
Because TopK selection is performed independently for each source node, $\mathbf{E}$ initially represents a directed graph. The independently estimated $E_{ij}$ and $E_{ji}$ need not be equal.

\begin{figure}[!t]
    \centering
    \includegraphics[width=\columnwidth]{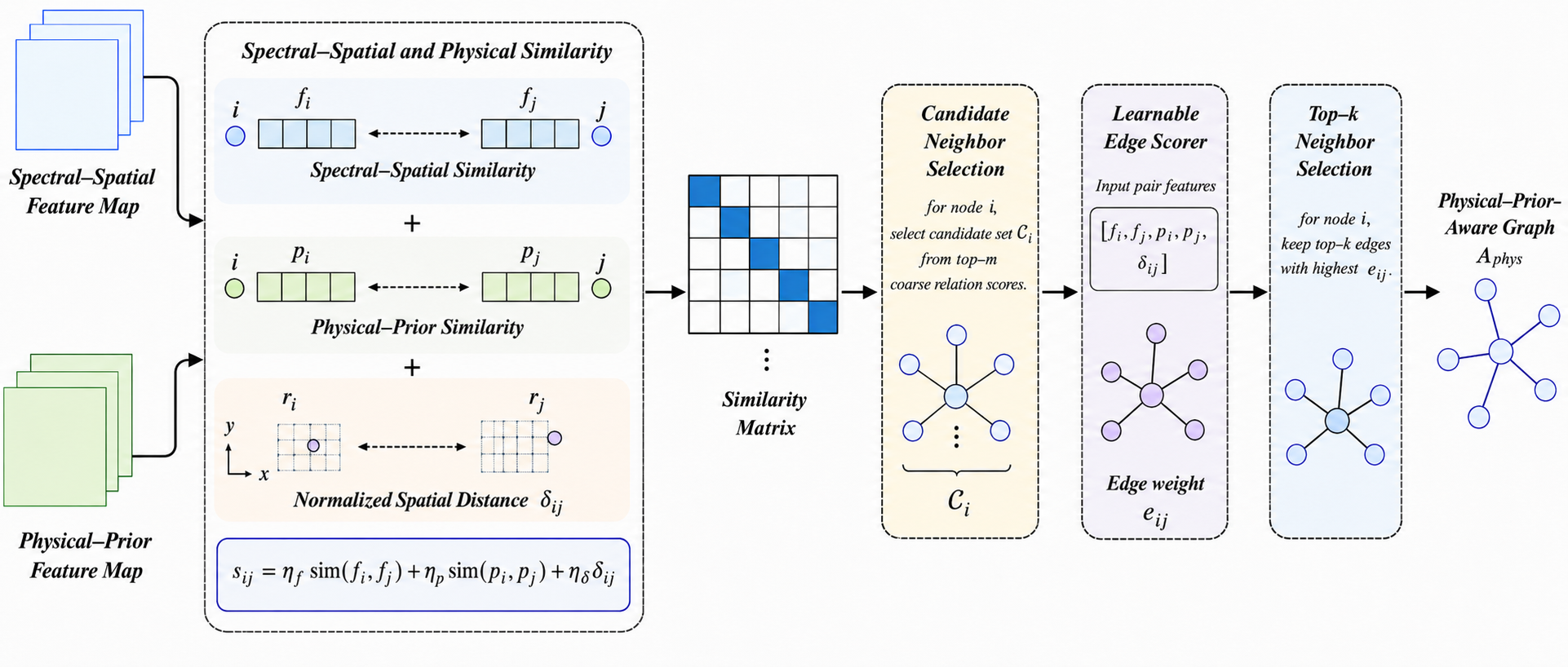}
    \caption{Two-stage physical-prior-aware adaptive graph construction.}
    \label{fig:adaptive_graph_construction}
\end{figure}

Consistent bidirectional connectivity is obtained by symmetrizing the directed edge-weight matrix:
\begin{equation}
\mathbf{A}
=
\frac{\mathbf{E}+\mathbf{E}^{\top}}{2}.
\label{eq:symmetric-adjacency}
\end{equation}
The weighted degree of node $i$ and the corresponding degree matrix are defined as
\begin{equation}
d_i
=
\sum_{j=1}^{N}A_{ij},
\label{eq:weighted-degree}
\end{equation}
\begin{equation}
\mathbf{D}
=
\operatorname{diag}(d_1,d_2,\ldots,d_N).
\label{eq:degree-matrix}
\end{equation}
The physical-prior-aware propagation matrix is then formulated as
\begin{equation}
\mathbf{A}_{\mathrm{phys}}
=
\frac{1}{2}
\left(
\mathbf{I}+\mathbf{D}^{-1}\mathbf{A}
\right)
\in\mathbb{R}^{N\times N}.
\label{eq:physical-propagation-matrix}
\end{equation}
Here, $\mathbf{D}^{-1}\mathbf{A}$ distributes row-normalized weights over neighboring nodes, $\mathbf{I}$ preserves self-information, and the factor $1/2$ balances self-preservation and neighborhood propagation while maintaining unit row sums. The nonzero entries of $\mathbf{A}_{\mathrm{phys}}$ determine the adaptive neighborhood of each node, and their values quantify the learned connection strengths. In the subsequent physics-gated graph diffusion module, $\mathbf{A}_{\mathrm{phys}}$ regulates the attention distribution and neighborhood feature propagation. Fig.~\ref{fig:adaptive_graph_construction} summarizes the two-stage construction process.

\subsection{Physics-Gated Graph Diffusion}
The physical-prior-aware graph provides the sparse topology and connection strengths for node-wise information propagation. At diffusion layer $\ell$, the current node state $\mathbf{H}^{(\ell)}\in\mathbb{R}^{N\times M}$ is first propagated over the retained graph connections and then adaptively fused with a layer-specific physical-prior representation. The resulting update is expressed as
\begin{equation}
\mathbf{H}^{(\ell+1)}
=
\mathbf{G}^{(\ell)}\odot\mathbf{Z}^{(\ell)}
+
\left(
\mathbf{1}-\mathbf{G}^{(\ell)}
\right)\odot\mathbf{P}^{(\ell)},
\label{eq:physics-gated-update}
\end{equation}
where $\mathbf{Z}^{(\ell)}\in\mathbb{R}^{N\times M}$ is the graph diffusion representation, $\mathbf{P}^{(\ell)}\in\mathbb{R}^{N\times M}$ is the physical prior projected into the current hidden space, and $\mathbf{G}^{(\ell)}\in(0,1)^{N\times M}$ is a node-wise and feature-wise gate. Here, $\mathbf{1}$ is an all-ones matrix and $\odot$ denotes elementwise multiplication. Before the first diffusion layer, the shallow node representation $\mathbf{F}$ is projected into the initial state $\mathbf{H}^{(0)}$.

The connection strengths in $\mathbf{A}_{\mathrm{phys}}$ provide persistent relation priors, and the current node states produce layer-dependent pairwise relevance. Let $\mathbf{B}^{(\ell)}\in\mathbb{R}^{N\times N}$ denote the data-driven attention score matrix derived from $\mathbf{H}^{(\ell)}$. The unnormalized edge-biased attention matrix is defined as
\begin{equation}
\mathbf{Q}^{(\ell)}
=
\mathbf{A}_{\mathrm{phys}}
\odot
\exp\left(\mathbf{B}^{(\ell)}\right).
\label{eq:edge-biased-attention-matrix}
\end{equation}
The exponential term converts the data-driven scores into positive relation weights, and $\mathbf{A}_{\mathrm{phys}}$ supplies the graph support and learned connection strengths. Its zero entries suppress information exchange between unconnected nodes. This multiplicative form is equivalent to adding the logarithm of each positive graph weight to the corresponding attention score. Row-wise normalization of $\mathbf{Q}^{(\ell)}$ yields the sparse propagation matrix $\mathbf{Q}_{\mathrm{norm}}^{(\ell)}\in\mathbb{R}^{N\times N}$.

The normalized propagation matrix aggregates the current node states over the retained sparse connections:
\begin{equation}
\mathbf{Z}^{(\ell)}
=
\mathbf{Q}_{\mathrm{norm}}^{(\ell)}
\mathbf{H}^{(\ell)}.
\label{eq:sparse-graph-diffusion}
\end{equation}
Each row of $\mathbf{Z}^{(\ell)}$ combines the states of the connected nodes according to the edge-biased attention distribution. In implementation, multihead value projections and head concatenation realize the sparse aggregation, followed by a residual projection, layer normalization, and nonlinear activation.

The encoder output $\mathbf{P}\in\mathbb{R}^{N\times M_p}$ is defined in the physical-prior space, and the graph diffusion representation $\mathbf{Z}^{(\ell)}\in\mathbb{R}^{N\times M}$ is defined in the hidden feature space. A layer-specific projector aligns the shared physical prior with the hidden space of diffusion layer $\ell$:
\begin{equation}
\mathbf{P}^{(\ell)}
=
\operatorname{Proj}^{(\ell)}
\left(
\mathbf{P}
\right).
\label{eq:layer-prior-projection}
\end{equation}
Here, $\operatorname{Proj}^{(\ell)}(\cdot)$ denotes the layer-specific prior transformation, and $\mathbf{P}^{(\ell)}\in\mathbb{R}^{N\times M}$ is the dimension-aligned physical-prior representation. Each diffusion layer derives $\mathbf{P}^{(\ell)}$ directly from the shared $\mathbf{P}$ for node-aligned gated fusion.

\begin{figure}[!t]
    \centering
    \includegraphics[width=\columnwidth]{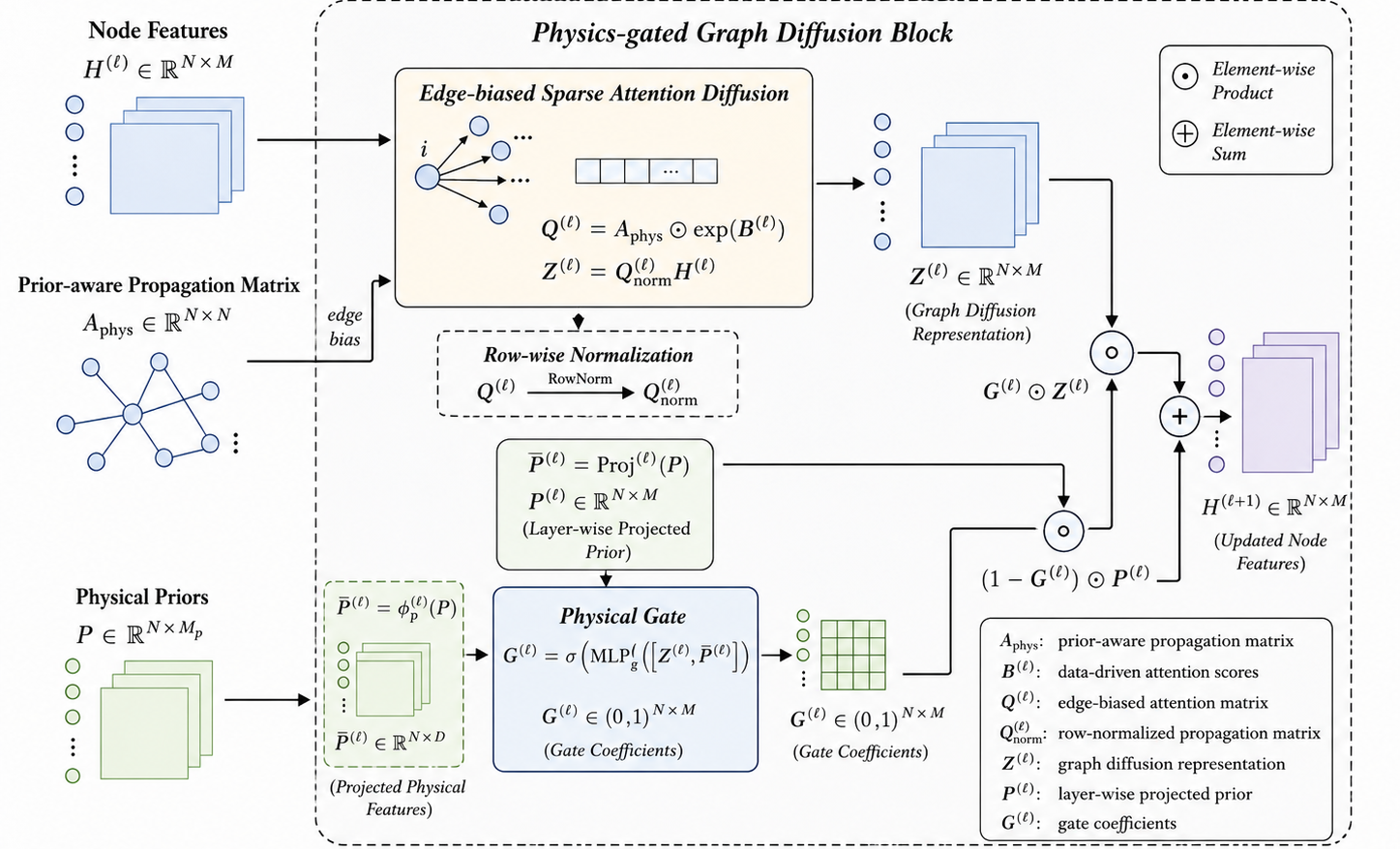}
    \caption{Edge-biased physics-gated graph diffusion in DAPGNet.}
    \label{fig:physical_graph_diffusion}
\end{figure}

The gate is jointly generated from the graph diffusion representation and the projected physical prior:
\begin{equation}
\mathbf{G}^{(\ell)}
=
\sigma\left(
\operatorname{MLP}
\left(
[\mathbf{Z}^{(\ell)},\mathbf{P}^{(\ell)}]
\right)
\right).
\label{eq:physical-gate}
\end{equation}
The bracket denotes concatenation along the feature dimension, and the MLP maps the resulting $N\times2M$ representation to an $N\times M$ gating matrix. Larger gate values increase the contribution of graph diffusion. Smaller gate values retain a greater proportion of the projected physical prior. The update in \eqref{eq:physics-gated-update} is applied for $\ell=0,1,\ldots,L-1$, and $\mathbf{H}^{(\ell+1)}$ becomes the input to the subsequent diffusion layer. Through this process, the learned graph weights regulate neighborhood propagation and the physical prior remains active in every layer. Fig.~\ref{fig:physical_graph_diffusion} illustrates the complete diffusion module.

\subsection{Cross-Layer Feature Fusion and Classification}
After the physics-gated graph diffusion stages, the network retains the initial graph state $\mathbf{H}^{(0)}$ and obtains the node representations $\mathbf{H}^{(1)},\ldots,\mathbf{H}^{(L)}$ at successive diffusion depths. The initial state preserves shallow spectral--spatial content, and the subsequent states encode graph dependencies accumulated through layer-wise diffusion. A cross-layer fusion module jointly integrates these complementary representations.

Let $\mathbf{H}^{(\ell)}\in\mathbb{R}^{N\times M}$, $\ell=0,1,\ldots,L$, denote the node representation at the $\ell$th diffusion depth. The module combines a concatenation-based nonlinear mapping with depth-adaptive weighted aggregation. The fused node representation is formulated as
\begin{equation}
\mathbf{H}_{\mathrm{f}}
=
\operatorname{MLP}_{\mathrm{f}}
\left(
[\mathbf{H}^{(0)},\mathbf{H}^{(1)},\ldots,\mathbf{H}^{(L)}]
\right)
+
\sum_{\ell=0}^{L}
\mathbf{W}^{(\ell)}\odot\mathbf{H}^{(\ell)}.
\label{eq:cross-layer-fusion}
\end{equation}
Here, $[\mathbf{H}^{(0)},\mathbf{H}^{(1)},\ldots,\mathbf{H}^{(L)}]\in\mathbb{R}^{N\times(L+1)M}$ denotes feature-wise concatenation, and $\operatorname{MLP}_{\mathrm{f}}(\cdot)$ maps the concatenated representation to $\mathbb{R}^{N\times M}$. The matrix $\mathbf{W}^{(\ell)}\in\mathbb{R}^{N\times1}$ contains the node-adaptive weights assigned to the $\ell$th representation and is broadcast along the feature dimension. These weights combine node-dependent scores with shared learnable depth preferences and are normalized across the $L+1$ diffusion depths. The nonlinear mapping captures joint interactions among representations from different diffusion depths. The weighted aggregation selectively retains the depth-specific information required by each node, yielding $\mathbf{H}_{\mathrm{f}}\in\mathbb{R}^{N\times M}$.

Average and maximum pooling summarize $\mathbf{H}_{\mathrm{f}}$ along the node dimension, and the average-pooled feature undergoes an additional transformation to produce an enhanced global representation. The two pooled descriptors and the enhanced global representation are concatenated and fed into the classifier to predict the category of the center pixel.
\subsection{Objective Function}
DAPGNet is optimized using a joint objective composed of the main classification loss, auxiliary classification loss, and spectral smoothness regularization.

Let $\widehat{\mathbf{y}}$ denote the final prediction of the center pixel and $y$ denote its ground-truth label. The main classification loss applies cross-entropy supervision to the final prediction:
\begin{equation}
\mathcal{L}_{\mathrm{cls}}
=
\operatorname{CE}
\left(
\widehat{\mathbf{y}},y
\right).
\label{eq:main-loss}
\end{equation}
Here, $\operatorname{CE}(\cdot,\cdot)$ denotes standard cross-entropy.

To provide direct supervision at different graph diffusion depths, lightweight auxiliary classifiers are applied to the intermediate node representations. Let $\widehat{\mathbf{y}}_{\mathrm{aux}}^{(\ell)}$ denote the auxiliary prediction generated at diffusion depth $\ell$. The auxiliary classification loss is defined as
\begin{equation}
\mathcal{L}_{\mathrm{aux}}
=
\frac{1}{L}
\sum_{\ell=1}^{L}
\operatorname{CE}
\left(
\widehat{\mathbf{y}}_{\mathrm{aux}}^{(\ell)},y
\right).
\label{eq:aux-loss}
\end{equation}
This loss averages the auxiliary supervision across the graph diffusion depths and promotes discriminative intermediate representations during training.

Spectral smoothness regularization is imposed on the branch responses before spectral pooling in the multiscale spectral physical prior encoder. Second-order spatio-spectral difference regularization has been used to characterize inherent spectral correlation and piecewise smoothness in HSI \cite{takemoto2025spatiospectral}. Let $\mathbf{u}_{t}^{(q)}$ denote the center-pixel response vector at spectral position $t$ from the branch whose initial kernel width is $q\in\{3,5,7\}$. The spectral smoothness term is formulated as
\begin{equation}
\mathcal{L}_{\mathrm{spe}}
=
\sum_{q\in\{3,5,7\}}
\sum_{t}
\left\lVert
\mathbf{u}_{t+1}^{(q)}
-2\mathbf{u}_{t}^{(q)}
+\mathbf{u}_{t-1}^{(q)}
\right\rVert_{2}^{2}.
\label{eq:spectral-loss}
\end{equation}
The summation covers the valid interior positions of each branch response sequence. This term penalizes abrupt second-order fluctuations along the ordered spectral dimension and encourages locally consistent spectral-response variation.

The complete objective is written as
\begin{equation}
\mathcal{L}
=
\mathcal{L}_{\mathrm{cls}}
+\lambda_{\mathrm{aux}}\mathcal{L}_{\mathrm{aux}}
+\lambda_{\mathrm{spe}}\mathcal{L}_{\mathrm{spe}}.
\label{eq:total-loss}
\end{equation}
Here, $\lambda_{\mathrm{aux}}$ and $\lambda_{\mathrm{spe}}$ control the contributions of auxiliary supervision and spectral smoothness regularization, respectively.
\section{Experiments}

\subsection{Datasets}
We conduct experiments on four public HSI benchmarks: Indian Pines, WHU-Hi-LongKou, Houston2013, and Houston2018.

\noindent\textbf{1) Indian Pines:} Indian Pines comes from an AVIRIS image collected over an agricultural area in northwestern Indiana, USA. The image size is \(145\times145\) pixels, with a spatial resolution of 20 m. The raw cube contains 224 spectral bands, and 200 effective bands are commonly retained after removing water-absorption and noisy bands. The ground truth contains 16 land-cover classes.

\noindent\textbf{2) WHU-Hi-LongKou:} WHU-Hi-LongKou is a UAV-borne hyperspectral scene from the WHU-Hi benchmark series. It was collected over LongKou, Hubei Province, China. The image size is \(550\times400\) pixels, with a ground sampling distance of 0.463 m. It contains 270 spectral bands and 9 land-cover classes.

\noindent\textbf{3) Houston2013:} Houston2013 comes from the 2013 IEEE GRSS Data Fusion Contest. It covers the University of Houston and its surrounding urban area. The hyperspectral image size is \(349\times1905\) pixels, with a spatial resolution of 2.5 m. It contains 144 spectral bands and 15 land-cover classes.

\noindent\textbf{4) Houston2018:} Houston2018 comes from the 2018 IEEE GRSS Data Fusion Contest. It covers an urban area in Houston, USA. The hyperspectral image used in this study has a size of \(601\times2384\) pixels, with a spatial resolution of 1 m. It contains 48 spectral bands and 20 land-cover classes.

\subsection{Experimental Details}
All experiments were implemented in PyTorch on a workstation equipped with an Intel Xeon Gold 6530 CPU and an NVIDIA GeForce RTX 4090 GPU. To reduce the influence of sample partitioning and parameter initialization, each experimental setting was repeated ten times with different random seeds. The mean and standard deviation are reported.

All models were optimized using AdamW with an initial learning rate of \tnrinline{1\times 10^{-3}} and a weight decay of \tnrinline{1\times 10^{-4}}. The learning rate was updated by a StepLR scheduler with a step size of 10 and a decay factor of 0.7. For the main comparative and ablation experiments, the spatial input patch size was fixed to \tnrinline{15\times 15}. For DAPGNet, the hidden dimension and physical-prior dimension were set to 128 and 64, respectively. The numbers of graph diffusion layers, adaptive neighbors, and attention heads were set to 2, 12, and 8, respectively.

\subsection{Evaluation Metrics}
We use overall accuracy (OA), average accuracy (AA), and the Kappa coefficient as evaluation metrics. OA measures the proportion of correctly classified test samples. AA averages the accuracy of each class and better reflects the performance on minority classes. Kappa measures the agreement between predicted and ground-truth labels after removing chance agreement.

\subsection{Comparative Experiments}
The comparative experiments evaluate whether introducing physical priors into graph-oriented HSI classification brings consistent advantages over representation-centered and relation-modeling baselines. DAPGNet is compared with representative CNN-, Transformer-, state-space/Mamba-, and graph-related classifiers, including 2D-CNN~\cite{liuSemisupervisedConvolutional2017}, 3D-CNN~\cite{benHamida3DDeepLearning2018}, HybridSN~\cite{royHybridSNExploring2020}, HIT~\cite{yangHyperspectralImageTransformer2022}, SSFTT~\cite{sunSpectralSpatialFeatureTokenization2022}, SS\_TMNET~\cite{huangSSTMNetSpatialSpectral2023}, MambaHSI~\cite{liMambaHSISpatialSpectral2024}, SSMamba~\cite{huangSpectralSpatialMamba2024}, ConvMamba~\cite{zhaoSpectralSelectionConvolution2026}, GraphMamba~\cite{yangGraphMambaEfficient2024}, DBMGNet~\cite{wangDBMGNetDualbranchMambaGCN2025}, and S2VNet~\cite{hanSubpixelSpectralVariability2025}. These baselines cover local convolutional spectral-spatial modeling, hybrid 2D/3D convolution, token-based dependency modeling, Mamba-style sequence modeling, and graph or graph-hybrid relation modeling.

Tables~\ref{tab:indianpines-01}--\ref{tab:houston2018-001} report the quantitative comparison on the four datasets. DAPGNet obtains the best OA, AA, and Kappa on all datasets, showing stable aggregate advantages across agricultural, urban, and UAV-borne scenes. The gains are particularly evident in AA, which suggests improved robustness to class imbalance and difficult categories rather than only better performance on dominant classes.

On Indian Pines, DAPGNet achieves $94.66\pm0.51\%$ OA, $83.73\pm3.79\%$ AA, and $93.90\pm0.58$ Kappa. Compared with CNN-, Transformer-, and Mamba-based competitors, the larger AA margin is consistent with the motivation of multiscale physical-prior encoding. Agricultural categories differ at multiple spectral variation scales, and explicitly using physical priors provides a more reliable basis for constructing pixel relations under limited supervision.

On WHU-Hi-LongKou, DAPGNet obtains $95.63\pm0.69\%$ OA, $82.69\pm3.49\%$ AA, and $94.21\pm0.93$ Kappa. This UAV-borne scene contains fine spatial details and uneven class distributions. Although several class-level best scores are achieved by other methods, DAPGNet produces the strongest aggregate results, indicating that the proposed framework remains robust when fine-grained crop textures and class imbalance appear together.

The two Houston benchmarks further test the method under heterogeneous urban layouts. On Houston2013, DAPGNet reaches $88.29\pm1.51\%$ OA and $88.60\pm1.45\%$ AA. On Houston2018, it obtains $91.63\pm0.19\%$ OA and $80.49\pm1.73\%$ AA, with a clear improvement in AA over graph-related baselines such as GraphMamba and DBMGNet. Urban scenes contain adjacent pixels with different materials and visually similar materials distributed across different structures. The results therefore support physical-prior-aware graph construction and physics-gated graph diffusion, where spectral--spatial affinity, physical-prior consistency, and learned edge weights jointly regulate neighborhood aggregation. Overall, the comparative results support the central claim that physical priors provide stronger support when they are coupled with node representation, pixel-relation modeling, and neighborhood diffusion.

\begin{figure*}[!t]
    \centering
    \includegraphics[width=\textwidth]{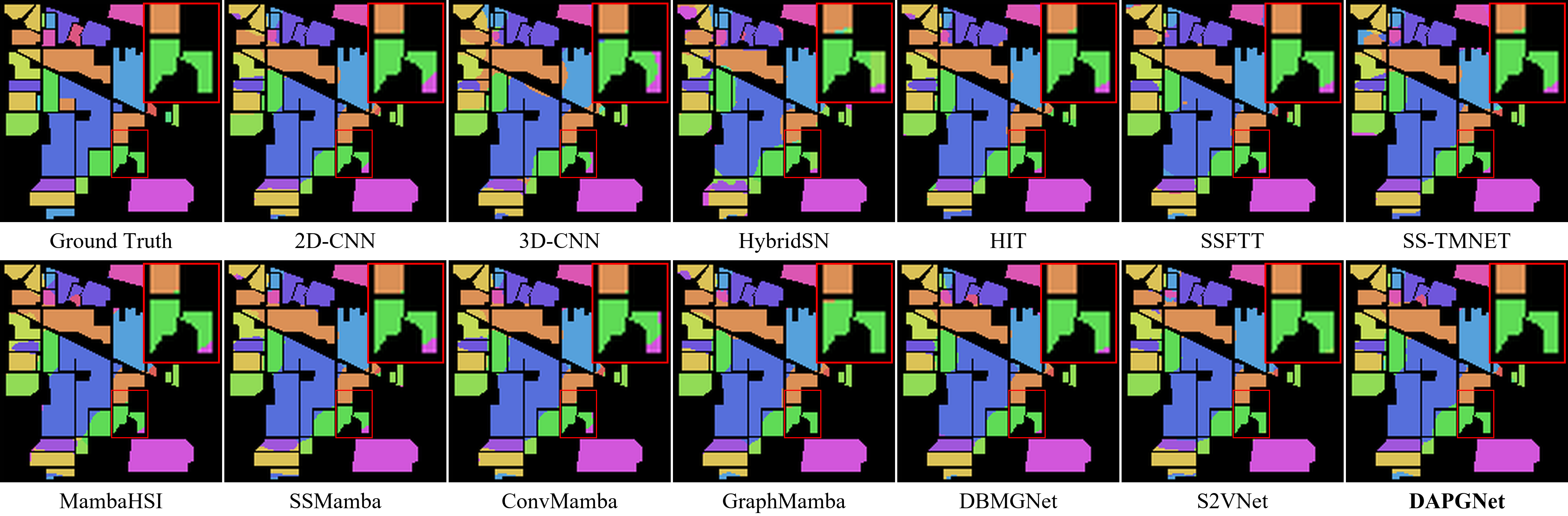}
    \caption{Classification maps on Indian Pines with 10\% training samples.}
    \label{fig:comparison-map-ip}
\end{figure*}

\begin{figure*}[!t]
    \centering
    \includegraphics[width=\textwidth]{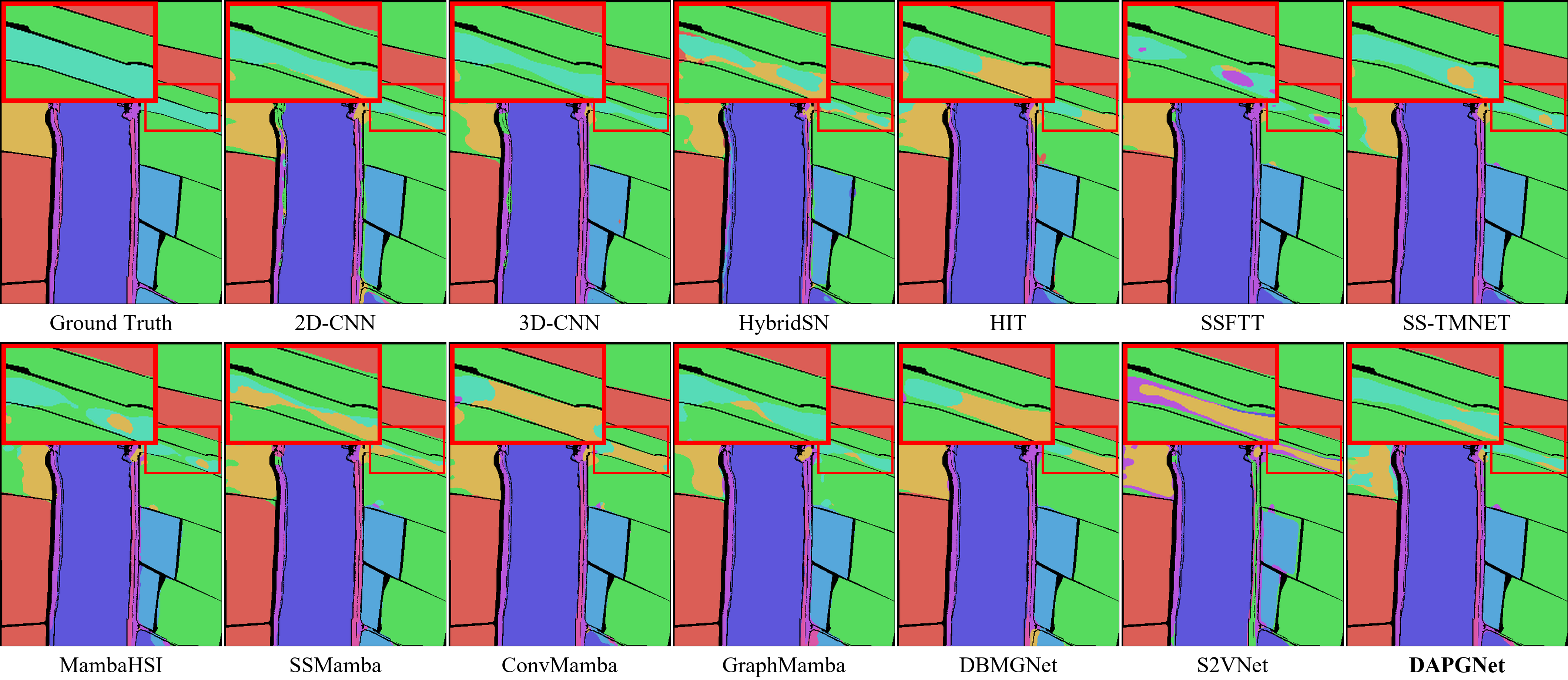}
    \caption{Classification maps on WHU-Hi-LongKou with 0.1\% training samples.}
    \label{fig:comparison-map-wh}
\end{figure*}

\begin{figure*}[!t]
    \centering
    \includegraphics[width=\textwidth]{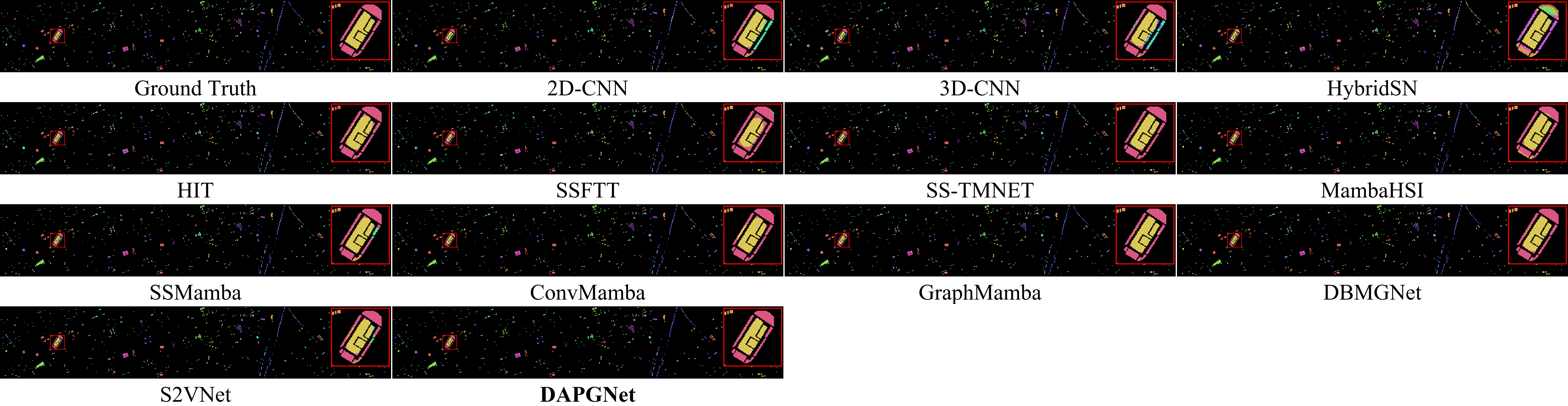}
    \caption{Classification maps on Houston2013 with 1\% training samples.}
    \label{fig:comparison-map-houston2013}
\end{figure*}

\begin{figure*}[!t]
    \centering
    \includegraphics[width=\textwidth]{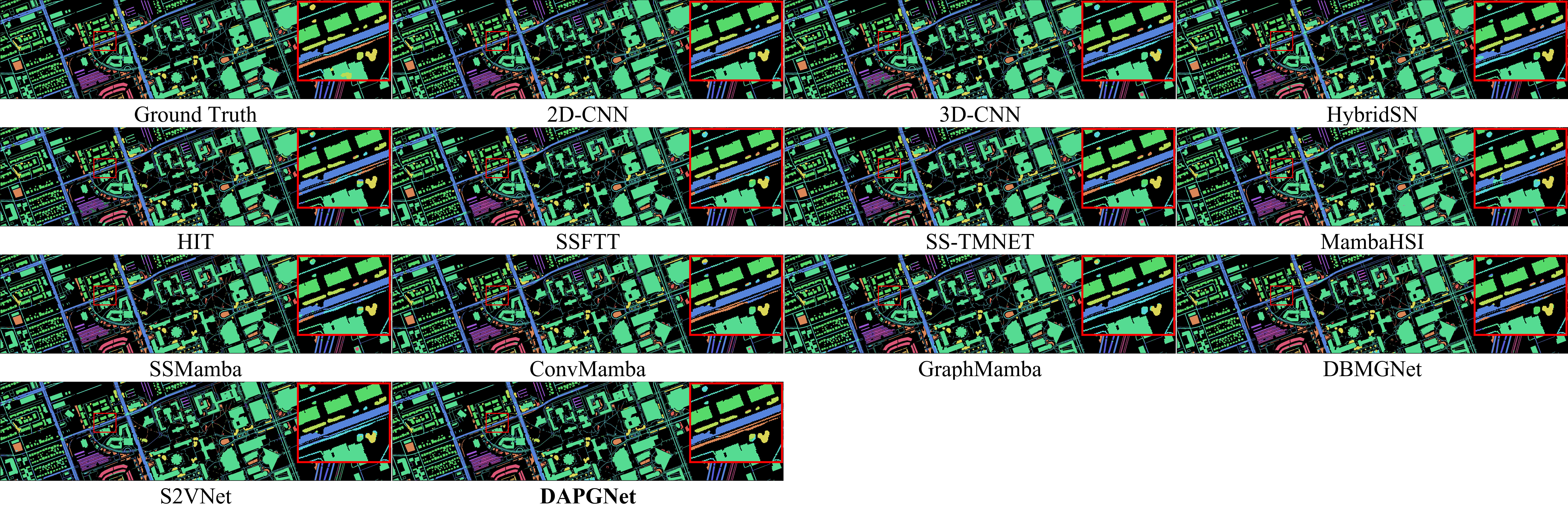}
    \caption{Classification maps on Houston2018 with 1\% training samples.}
    \label{fig:comparison-map-houston2018}
\end{figure*}

Figs.~\ref{fig:comparison-map-ip}--\ref{fig:comparison-map-houston2018} provide qualitative comparisons of the classification maps across the four datasets. The visual results complement the quantitative comparisons by showing prediction consistency in agricultural, UAV-borne, and urban scenes. DAPGNet generally produces spatially more coherent predictions in heterogeneous regions.

\begin{table*}[!t]
    \centering
    \caption{Classification results on Indian Pines (10\% training samples)}
    \label{tab:indianpines-01}
    \begingroup
    \tiny\rmfamily
    \setlength{\tabcolsep}{3pt}
    \renewcommand{\arraystretch}{1.12}
    \resizebox{\textwidth}{!}{%
    \begin{tabular}{lcccccccccccccc}
        \toprule
        Class & 2D-CNN & 3D-CNN & HybridSN & HIT & SSFTT & SS\_TMNET & MambaHSI & SSMamba & ConvMamba & GraphMamba & DBMGNet & S2VNet & DAPGNet \\
        \midrule
        Alfalfa & $30.71 \pm 24.52$ & $27.38 \pm 26.54$ & $0.00 \pm 0.00$ & $7.62 \pm 9.75$ & $41.43 \pm 34.51$ & $41.43 \pm 27.77$ & $57.86 \pm 17.95$ & $28.10 \pm 21.16$ & $36.43 \pm 23.04$  & $46.82 \pm 21.72$ & $62.11 \pm 20.95$ & $33.10 \pm 26.07$ & $\mathbf{76.19 \pm 27.09}$ \\
        Corn-notill & $91.19 \pm 1.03$ & $89.18 \pm 2.32$ & $72.65 \pm 20.12$ & $89.59 \pm 1.93$ & $92.84 \pm 1.33$ & $84.32 \pm 6.17$ & $89.42 \pm 1.31$ & $91.98 \pm 2.66$ & $91.96 \pm 1.91$  & $92.23 \pm 1.20$ & $91.77 \pm 2.55$ & $\mathbf{93.24 \pm 1.32}$ & $90.79 \pm 1.15$ \\
        Corn-mintill & $96.36 \pm 1.59$ & $79.69 \pm 6.13$ & $60.03 \pm 24.59$ & $90.21 \pm 3.42$ & $95.35 \pm 3.78$ & $91.70 \pm 3.83$ & $89.91 \pm 3.31$ & $89.71 \pm 2.95$ & $95.93 \pm 2.28$  & $94.81 \pm 3.05$ & $\mathbf{97.19 \pm 1.19}$ & $93.65 \pm 2.81$ & $96.83 \pm 2.58$ \\
        Corn & $88.50 \pm 6.19$ & $71.12 \pm 9.58$ & $58.69 \pm 29.03$ & $87.85 \pm 8.43$ & $93.32 \pm 7.31$ & $83.18 \pm 6.77$ & $97.10 \pm 2.76$ & $92.10 \pm 3.33$ & $92.52 \pm 9.34$  & $\mathbf{97.55 \pm 3.81}$ & $95.44 \pm 4.73$ & $86.96 \pm 13.38$ & $96.07 \pm 5.35$ \\
        Grass-pasture & $91.10 \pm 4.19$ & $88.32 \pm 3.61$ & $71.59 \pm 34.71$ & $83.36 \pm 4.96$ & $93.45 \pm 3.27$ & $87.63 \pm 4.98$ & $86.39 \pm 7.10$ & $\mathbf{94.51 \pm 0.94}$ & $93.45 \pm 3.48$  & $93.39 \pm 2.05$ & $93.91 \pm 2.24$ & $93.28 \pm 2.20$ & $94.23 \pm 1.96$ \\
        Grass-trees & $93.29 \pm 2.15$ & $91.58 \pm 2.33$ & $80.21 \pm 26.96$ & $94.84 \pm 1.04$ & $95.04 \pm 2.01$ & $93.82 \pm 1.75$ & $92.85 \pm 2.65$ & $93.53 \pm 1.53$ & $92.88 \pm 1.96$  & $94.39 \pm 1.80$ & $94.86 \pm 2.05$ & $94.11 \pm 2.29$ & $\mathbf{95.46 \pm 1.90}$ \\
        Grass-pasture-mowed & $0.00 \pm 0.00$ & $0.00 \pm 0.00$ & $0.00 \pm 0.00$ & $0.00 \pm 0.00$ & $12.69 \pm 23.71$ & $0.00 \pm 0.00$ & $0.00 \pm 0.00$ & $0.00 \pm 0.00$ & $0.00 \pm 0.00$  & $93.39 \pm 2.05$ & $\mathbf{93.91 \pm 2.24}$ & $93.28 \pm 2.20$ & $29.62 \pm 30.92$ \\
        Hay-windrowed & $99.40 \pm 1.25$ & $97.40 \pm 2.65$ & $86.94 \pm 29.13$ & $\mathbf{100.00 \pm 0.00}$ & $\mathbf{100.00 \pm 0.00}$ & $\mathbf{100.00 \pm 0.00}$ & $99.98 \pm 0.07$ & $99.95 \pm 0.14$ & $\mathbf{100.00 \pm 0.00}$ & $\mathbf{100.00 \pm 0.00}$ & $\mathbf{100.00 \pm 0.00}$ & $99.05 \pm 1.80$ & $99.98 \pm 0.07$ \\
        Oats & $0.00 \pm 0.00$ & $0.00 \pm 0.00$ & $0.00 \pm 0.00$ & $0.00 \pm 0.00$ & $0.00 \pm 0.00$ & $0.00 \pm 0.00$ & $0.56 \pm 1.67$ & $0.00 \pm 0.00$ & $0.00 \pm 0.00$  & $0.00 \pm 0.00$ & $0.46 \pm 1.54$ & $8.89 \pm 17.95$ & $\mathbf{39.44 \pm 31.86}$ \\
        Soybean-notill & $79.04 \pm 1.79$ & $70.66 \pm 3.52$ & $66.03 \pm 23.03$ & $75.94 \pm 3.05$ & $81.84 \pm 3.07$ & $76.25 \pm 2.88$ & $80.99 \pm 1.99$ & $81.76 \pm 1.56$ & $81.31 \pm 2.74$  & $84.46 \pm 2.63$ & $84.23 \pm 4.21$ & $83.11 \pm 2.61$ & $\mathbf{88.21 \pm 2.19}$ \\
        Soybean-mintill & $98.15 \pm 0.70$ & $93.13 \pm 3.06$ & $90.87 \pm 1.93$ & $96.08 \pm 1.54$ & $\mathbf{98.73 \pm 0.53}$ & $96.01 \pm 1.01$ & $97.41 \pm 1.16$ & $97.83 \pm 0.62$ & $98.01 \pm 0.80$  & $97.82 \pm 0.97$ & $98.28 \pm 0.49$ & $97.52 \pm 1.23$ & $98.04 \pm 0.73$ \\
        Soybean-clean & $94.64 \pm 2.10$ & $85.69 \pm 5.32$ & $69.01 \pm 27.46$ & $89.55 \pm 4.93$ & $93.50 \pm 2.84$ & $92.60 \pm 3.28$ & $94.34 \pm 2.91$ & $93.69 \pm 3.15$ & $94.64 \pm 3.12$  & $89.76 \pm 6.33$ & $\mathbf{94.71 \pm 2.31}$ & $92.90 \pm 2.54$ & $94.03 \pm 2.63$ \\
        Wheat & $82.76 \pm 5.56$ & $82.49 \pm 8.63$ & $36.49 \pm 34.68$ & $88.97 \pm 15.55$ & $88.27 \pm 5.79$ & $95.68 \pm 6.20$ & $88.70 \pm 19.00$ & $79.62 \pm 11.19$ & $90.70 \pm 5.95$  & $94.64 \pm 4.53$ & $\mathbf{96.85 \pm 2.43}$ & $92.76 \pm 6.74$ & $96.16 \pm 3.69$ \\
        Woods & $99.25 \pm 0.37$ & $98.28 \pm 1.18$ & $98.68 \pm 1.52$ & $99.18 \pm 0.43$ & $99.51 \pm 0.46$ & $99.69 \pm 0.24$ & $99.58 \pm 0.23$ & $99.66 \pm 0.27$ & $\mathbf{99.76 \pm 0.17}$ & $99.51 \pm 0.40$ & $99.66 \pm 0.23$ & $98.70 \pm 1.21$ & $99.44 \pm 0.54$ \\
        Buildings-Grass-Trees-Drives & $88.48 \pm 4.58$ & $85.63 \pm 4.59$ & $49.71 \pm 32.27$ & $87.70 \pm 5.61$ & $86.90 \pm 6.51$ & $86.06 \pm 6.46$ & $90.78 \pm 3.72$ & $85.52 \pm 8.27$ & $94.14 \pm 2.56$  & $90.28 \pm 5.96$ & $94.54 \pm 4.07$ & $83.13 \pm 8.68$ & $\mathbf{96.01 \pm 2.74}$ \\
        Stone-Steel-Towers & $31.90 \pm 24.30$ & $5.83 \pm 7.69$ & $2.74 \pm 4.33$ & $17.74 \pm 18.12$ & $22.26 \pm 19.12$ & $15.83 \pm 18.02$ & $27.02 \pm 20.60$ & $10.95 \pm 12.25$ & $34.64 \pm 16.06$  & $28.77 \pm 29.32$ & $36.91 \pm 21.64$ & $18.45 \pm 16.48$ & $\mathbf{49.16 \pm 21.37}$ \\
        \midrule
        OA (\%) & $92.26 \pm 0.63$ & $86.89 \pm 1.10$ & $76.29 \pm 14.25$ & $90.07 \pm 1.07$ & $93.20 \pm 0.76$ & $89.94 \pm 1.01$ & $91.79 \pm 0.75$ & $91.92 \pm 0.97$ & $93.21 \pm 0.84$  & $93.05 \pm 0.41$ & $93.98 \pm 0.46$ & $92.44 \pm 0.79$ & $\mathbf{94.66 \pm 0.51}$ \\
        AA (\%) & $72.80 \pm 3.03$ & $66.65 \pm 2.60$ & $52.73 \pm 14.95$ & $69.29 \pm 2.35$ & $74.70 \pm 3.91$ & $71.51 \pm 2.81$ & $74.56 \pm 2.02$ & $71.18 \pm 2.51$ & $74.77 \pm 2.64$  & $74.73 \pm 2.68$ & $77.80 \pm 2.25$ & $73.75 \pm 2.99$ & $\mathbf{83.73 \pm 3.79}$ \\
        Kappa & $91.15 \pm 0.73$ & $84.97 \pm 1.26$ & $72.28 \pm 17.78$ & $88.63 \pm 1.24$ & $92.23 \pm 0.87$ & $88.49 \pm 1.15$ & $90.60 \pm 0.86$ & $90.75 \pm 1.12$ & $92.23 \pm 0.96$  & $92.05 \pm 0.47$ & $93.12 \pm 0.53$ & $91.35 \pm 0.91$ & $\mathbf{93.90 \pm 0.58}$ \\
        \bottomrule
    \end{tabular}%
    }
    \endgroup
\end{table*}

\begin{table*}[!t]
    \centering
    \caption{Classification results on WHU-Hi-LongKou (0.1\% training samples)}
    \label{tab:wh-dapg}
    \begingroup
    \tiny\rmfamily
    \setlength{\tabcolsep}{3pt}
    \renewcommand{\arraystretch}{1.12}
    \resizebox{\textwidth}{!}{%
    \begin{tabular}{lcccccccccccccc}
        \toprule
        Class & 2D-CNN & 3D-CNN & HybridSN & HIT & SSFTT & SS\_TMNET & MambaHSI & SSMamba & ConvMamba & GraphMamba & DBMGNet & S2VNet & DAPGNet \\
        \midrule
        Corn & $98.15 \pm 0.90$ & $96.76 \pm 0.70$ & $97.36 \pm 1.28$ & $99.49 \pm 0.33$ & $99.09 \pm 0.59$ & $99.21 \pm 0.23$ & $98.91 \pm 0.78$ & $96.33 \pm 1.57$ & $99.81 \pm 0.15$  & $99.35 \pm 0.68$ & $\mathbf{99.88 \pm 0.14}$ & $95.19 \pm 13.01$ & $99.70 \pm 0.26$ \\
        Cotton & $96.27 \pm 2.84$ & $91.18 \pm 9.69$ & $69.95 \pm 28.23$ & $86.02 \pm 9.96$ & $73.37 \pm 25.59$ & $88.50 \pm 9.93$ & $90.15 \pm 11.87$ & $85.53 \pm 8.22$ & $93.99 \pm 4.37$  & $81.37 \pm 13.01$ & $\mathbf{96.56 \pm 1.13}$ & $43.96 \pm 41.33$ & $90.71 \pm 6.05$ \\
        Sesame & $41.30 \pm 26.21$ & $6.13 \pm 8.45$ & $0.27 \pm 0.44$ & $3.87 \pm 5.13$ & $0.00 \pm 0.00$ & $4.39 \pm 12.75$ & $22.42 \pm 21.81$ & $0.00 \pm 0.00$ & $64.76 \pm 26.54$  & $62.98 \pm 29.30$ & $59.34 \pm 34.38$ & $0.10 \pm 0.21$ & $\mathbf{76.28 \pm 14.01}$ \\
        Broad-leaf soybean & $99.28 \pm 0.42$ & $98.57 \pm 0.76$ & $98.32 \pm 1.44$ & $97.79 \pm 1.32$ & $98.74 \pm 0.72$ & $98.95 \pm 0.46$ & $97.83 \pm 1.20$ & $98.81 \pm 0.92$ & $98.98 \pm 0.46$  & $97.21 \pm 1.74$ & $99.30 \pm 0.40$ & $\mathbf{99.65 \pm 0.27}$ & $98.99 \pm 0.72$ \\
        Narrow-leaf soybean & $52.54 \pm 15.52$ & $\mathbf{60.01 \pm 13.46}$ & $35.02 \pm 22.43$ & $23.36 \pm 13.78$ & $20.14 \pm 27.86$ & $41.08 \pm 21.57$ & $41.21 \pm 15.39$ & $17.21 \pm 14.80$ & $40.28 \pm 13.86$  & $30.09 \pm 18.56$ & $47.12 \pm 19.60$ & $1.17 \pm 1.69$ & $58.03 \pm 14.88$ \\
        Rice & $89.27 \pm 4.04$ & $92.63 \pm 4.43$ & $85.26 \pm 17.28$ & $93.13 \pm 4.99$ & $92.97 \pm 5.66$ & $94.37 \pm 3.06$ & $92.78 \pm 2.98$ & $94.23 \pm 4.93$ & $94.83 \pm 3.50$  & $\mathbf{97.13 \pm 1.96}$ & $95.82 \pm 3.59$ & $45.84 \pm 44.29$ & $96.87 \pm 3.18$ \\
        Water & $99.90 \pm 0.11$ & $99.33 \pm 0.97$ & $\mathbf{100.00 \pm 0.00}$ & $\mathbf{100.00 \pm 0.00}$ & $\mathbf{100.00 \pm 0.01}$ & $\mathbf{100.00 \pm 0.00}$ & $99.98 \pm 0.05$ & $99.97 \pm 0.03$ & $99.99 \pm 0.01$  & $\mathbf{100.00 \pm 0.01}$ & $\mathbf{100.00 \pm 0.00}$ & $99.98 \pm 0.04$ & $\mathbf{100.00 \pm 0.01}$ \\
        Roads and houses & $70.93 \pm 9.75$ & $58.44 \pm 11.54$ & $42.96 \pm 17.95$ & $\mathbf{74.86 \pm 4.07}$ & $66.35 \pm 14.24$ & $70.65 \pm 10.82$ & $72.77 \pm 9.22$ & $66.74 \pm 13.14$ & $70.66 \pm 13.64$  & $69.96 \pm 8.30$ & $71.77 \pm 12.08$ & $23.10 \pm 28.48$ & $68.90 \pm 11.11$ \\
        Mixed weed & $38.15 \pm 24.06$ & $\mathbf{64.09 \pm 15.54}$ & $13.05 \pm 12.71$ & $32.11 \pm 16.01$ & $33.97 \pm 20.93$ & $46.74 \pm 12.78$ & $44.46 \pm 20.13$ & $36.91 \pm 27.92$ & $31.80 \pm 22.33$  & $44.37 \pm 17.19$ & $48.11 \pm 14.35$ & $5.71 \pm 10.09$ & $54.70 \pm 14.22$ \\
        \midrule
        OA (\%) & $94.23 \pm 0.62$ & $93.44 \pm 0.95$ & $89.94 \pm 2.21$ & $92.67 \pm 0.62$ & $92.00 \pm 1.70$ & $93.75 \pm 0.67$ & $93.61 \pm 0.87$ & $92.14 \pm 0.64$ & $94.60 \pm 0.77$  & $93.38 \pm 1.24$ & $95.23 \pm 0.45$ & $85.07 \pm 4.76$ & $\mathbf{95.63 \pm 0.69}$ \\
        AA (\%) & $76.20 \pm 3.78$ & $74.13 \pm 3.74$ & $60.24 \pm 6.69$ & $67.85 \pm 1.97$ & $64.96 \pm 6.01$ & $71.54 \pm 2.67$ & $73.39 \pm 4.03$ & $66.19 \pm 2.77$ & $77.23 \pm 3.84$  & $74.82 \pm 5.05$ & $79.05 \pm 3.61$ & $46.08 \pm 8.67$ & $\mathbf{82.69 \pm 3.49}$ \\
        Kappa & $92.32 \pm 0.84$ & $91.26 \pm 1.29$ & $86.43 \pm 3.05$ & $90.23 \pm 0.83$ & $89.28 \pm 2.32$ & $91.66 \pm 0.90$ & $91.49 \pm 1.17$ & $89.49 \pm 0.89$ & $92.83 \pm 1.03$  & $91.21 \pm 1.63$ & $93.67 \pm 0.61$ & $79.44 \pm 7.00$ & $\mathbf{94.21 \pm 0.93}$ \\
        \bottomrule
    \end{tabular}%
    }
    \endgroup
\end{table*}

\begin{table*}[!t]
    \centering
    \caption{Classification results on Houston2013 (1\% training samples)}
    \label{tab:houston2013-001}
    \begingroup
    \tiny\rmfamily
    \setlength{\tabcolsep}{3pt}
    \renewcommand{\arraystretch}{1.12}
    \resizebox{\textwidth}{!}{%
    \begin{tabular}{lcccccccccccccc}
        \toprule
        Class & 2D-CNN & 3D-CNN & HybridSN & HIT & SSFTT & SS\_TMNET & MambaHSI & SSMamba & ConvMamba & GraphMamba & DBMGNet & S2VNet & DAPGNet \\
        \midrule
        Healthy grass & $84.91 \pm 3.76$ & $74.97 \pm 7.00$ & $78.80 \pm 4.70$ & $77.13 \pm 1.98$ & $\mathbf{86.76 \pm 4.97}$ & $79.84 \pm 2.58$ & $81.02 \pm 4.66$ & $85.04 \pm 3.62$ & $86.12 \pm 4.40$  & $82.66 \pm 4.69$ & $84.95 \pm 3.33$ & $85.83 \pm 4.45$ & $86.30 \pm 5.06$ \\
        Stressed grass & $90.06 \pm 5.57$ & $81.55 \pm 6.21$ & $63.34 \pm 22.09$ & $82.79 \pm 8.98$ & $74.07 \pm 13.46$ & $80.14 \pm 9.51$ & $88.85 \pm 4.63$ & $91.14 \pm 2.11$ & $90.99 \pm 6.06$  & $87.58 \pm 5.66$ & $90.80 \pm 2.28$ & $91.02 \pm 5.52$ & $\mathbf{92.14 \pm 5.01}$ \\
        Synthetic grass & $89.38 \pm 7.46$ & $75.67 \pm 11.73$ & $84.01 \pm 13.81$ & $94.10 \pm 2.55$ & $74.44 \pm 37.44$ & $96.45 \pm 1.61$ & $89.68 \pm 6.53$ & $88.34 \pm 16.99$ & $97.47 \pm 1.40$  & $96.98 \pm 1.88$ & $94.62 \pm 2.71$ & $91.22 \pm 5.08$ & $\mathbf{97.52 \pm 1.56}$ \\
        Trees & $87.35 \pm 7.32$ & $87.27 \pm 7.44$ & $\mathbf{94.90 \pm 3.27}$ & $62.25 \pm 15.05$ & $87.92 \pm 7.27$ & $80.50 \pm 9.00$ & $86.78 \pm 3.65$ & $88.38 \pm 2.52$ & $87.38 \pm 7.86$  & $79.32 \pm 4.16$ & $89.51 \pm 4.91$ & $88.58 \pm 5.86$ & $89.01 \pm 4.28$ \\
        Soil & $96.77 \pm 3.63$ & $85.65 \pm 9.58$ & $72.26 \pm 27.90$ & $97.78 \pm 5.29$ & $94.01 \pm 5.52$ & $97.01 \pm 5.18$ & $98.01 \pm 2.28$ & $99.06 \pm 0.63$ & $\mathbf{99.37 \pm 1.11}$ & $97.38 \pm 3.24$ & $98.92 \pm 3.23$ & $97.11 \pm 5.04$ & $96.51 \pm 4.28$ \\
        Water & $54.41 \pm 16.17$ & $45.62 \pm 13.78$ & $16.55 \pm 24.06$ & $36.02 \pm 15.05$ & $18.85 \pm 19.59$ & $45.53 \pm 22.53$ & $60.93 \pm 16.15$ & $48.45 \pm 15.37$ & $58.14 \pm 18.45$  & $66.33 \pm 10.42$ & $74.28 \pm 2.37$ & $28.63 \pm 24.45$ & $\mathbf{79.35 \pm 5.57}$ \\
        Residential & $69.44 \pm 14.05$ & $78.04 \pm 11.23$ & $48.74 \pm 25.18$ & $63.12 \pm 11.34$ & $86.22 \pm 11.36$ & $86.25 \pm 6.55$ & $80.84 \pm 9.32$ & $45.55 \pm 16.58$ & $51.30 \pm 22.45$  & $75.56 \pm 9.70$ & $87.83 \pm 7.57$ & $71.46 \pm 16.45$ & $\mathbf{89.24 \pm 4.34}$ \\
        Commercial & $67.35 \pm 6.28$ & $64.67 \pm 6.70$ & $43.14 \pm 13.78$ & $57.89 \pm 6.56$ & $47.65 \pm 11.99$ & $60.86 \pm 6.99$ & $\mathbf{71.70 \pm 4.38}$ & $71.17 \pm 7.54$ & $70.82 \pm 7.25$  & $69.43 \pm 7.60$ & $68.60 \pm 9.58$ & $65.72 \pm 10.81$ & $64.89 \pm 9.40$ \\
        Road & $79.89 \pm 7.33$ & $78.50 \pm 5.79$ & $28.30 \pm 24.14$ & $71.55 \pm 5.34$ & $80.93 \pm 6.17$ & $79.05 \pm 5.23$ & $87.15 \pm 5.48$ & $63.90 \pm 7.28$ & $69.19 \pm 6.43$  & $82.96 \pm 6.35$ & $86.13 \pm 4.68$ & $83.60 \pm 5.60$ & $\mathbf{87.34 \pm 5.60}$ \\
        Highway & $87.39 \pm 6.36$ & $73.16 \pm 7.42$ & $37.35 \pm 24.37$ & $78.35 \pm 8.40$ & $56.24 \pm 20.37$ & $85.07 \pm 8.95$ & $89.78 \pm 6.52$ & $85.22 \pm 8.84$ & $89.14 \pm 10.69$  & $86.71 \pm 6.38$ & $87.98 \pm 8.41$ & $84.89 \pm 10.50$ & $\mathbf{96.21 \pm 3.61}$ \\
        Railway & $75.27 \pm 14.63$ & $79.01 \pm 8.67$ & $31.89 \pm 22.83$ & $35.04 \pm 15.13$ & $75.27 \pm 19.86$ & $79.96 \pm 8.01$ & $86.67 \pm 5.22$ & $67.09 \pm 14.53$ & $69.23 \pm 20.44$  & $81.99 \pm 8.25$ & $89.66 \pm 7.06$ & $78.72 \pm 10.25$ & $\mathbf{91.45 \pm 5.95}$ \\
        Parking Lot 1 & $82.10 \pm 9.51$ & $66.25 \pm 11.40$ & $55.81 \pm 17.56$ & $75.16 \pm 7.54$ & $68.48 \pm 18.08$ & $72.24 \pm 8.05$ & $70.90 \pm 8.72$ & $76.31 \pm 9.21$ & $\mathbf{84.74 \pm 7.26}$ & $77.41 \pm 8.61$ & $75.58 \pm 8.35$ & $79.48 \pm 9.47$ & $81.39 \pm 8.65$ \\
        Parking Lot 2 & $77.61 \pm 14.59$ & $66.34 \pm 13.74$ & $63.16 \pm 33.98$ & $19.48 \pm 22.00$ & $49.70 \pm 29.77$ & $37.55 \pm 22.25$ & $69.70 \pm 11.01$ & $13.08 \pm 18.02$ & $11.74 \pm 15.62$  & $49.30 \pm 27.43$ & $53.91 \pm 24.13$ & $8.80 \pm 15.72$ & $\mathbf{83.36 \pm 6.80}$ \\
        Tennis Court & $88.37 \pm 6.88$ & $82.08 \pm 9.75$ & $39.98 \pm 21.61$ & $86.04 \pm 9.21$ & $60.19 \pm 28.19$ & $84.88 \pm 14.03$ & $95.92 \pm 3.11$ & $76.79 \pm 20.96$ & $98.23 \pm 1.61$  & $89.74 \pm 4.72$ & $97.46 \pm 3.95$ & $86.84 \pm 12.03$ & $\mathbf{98.94 \pm 0.99}$ \\
        Running Track & $75.50 \pm 11.39$ & $70.69 \pm 9.50$ & $16.82 \pm 14.70$ & $82.95 \pm 13.87$ & $67.26 \pm 24.87$ & $92.66 \pm 7.24$ & $90.54 \pm 5.37$ & $83.88 \pm 12.51$ & $84.13 \pm 10.57$  & $88.39 \pm 7.42$ & $\mathbf{97.26 \pm 4.12}$ & $79.79 \pm 20.62$ & $95.44 \pm 5.14$ \\
        \midrule
        OA (\%) & $81.53 \pm 1.76$ & $75.73 \pm 2.02$ & $54.06 \pm 10.28$ & $69.93 \pm 2.55$ & $72.88 \pm 6.66$ & $79.47 \pm 3.28$ & $84.08 \pm 1.19$ & $75.39 \pm 2.99$ & $78.71 \pm 4.10$  & $81.42 \pm 1.70$ & $85.34 \pm 1.96$ & $79.55 \pm 2.58$ & $\mathbf{88.29 \pm 1.51}$ \\
        AA (\%) & $80.39 \pm 1.48$ & $73.97 \pm 2.03$ & $51.67 \pm 10.83$ & $67.98 \pm 2.68$ & $68.53 \pm 8.56$ & $77.20 \pm 3.72$ & $83.23 \pm 1.52$ & $72.23 \pm 4.02$ & $76.53 \pm 4.36$  & $80.13 \pm 2.87$ & $84.57 \pm 2.09$ & $74.78 \pm 2.79$ & $\mathbf{88.60 \pm 1.45}$ \\
        Kappa & $80.01 \pm 1.90$ & $73.72 \pm 2.19$ & $50.24 \pm 11.24$ & $67.42 \pm 2.76$ & $70.58 \pm 7.28$ & $77.76 \pm 3.56$ & $82.78 \pm 1.29$ & $73.33 \pm 3.25$ & $76.94 \pm 4.45$  & $79.89 \pm 1.85$ & $84.13 \pm 2.12$ & $77.83 \pm 2.80$ & $\mathbf{87.33 \pm 1.63}$ \\
        \bottomrule
    \end{tabular}%
    }
    \endgroup
\end{table*}

\begin{table*}[!t]
    \centering
    \caption{Classification results on Houston2018 (1\% training samples)}
    \label{tab:houston2018-001}
    \begingroup
    \tiny\rmfamily
    \setlength{\tabcolsep}{3pt}
    \renewcommand{\arraystretch}{1.12}
    \resizebox{\textwidth}{!}{%
    \begin{tabular}{lcccccccccccccc}
        \toprule
        Class & 2D-CNN & 3D-CNN & HybridSN & HIT & SSFTT & SS\_TMNET & MambaHSI & SSMamba & ConvMamba & GraphMamba & DBMGNet & S2VNet & DAPGNet \\
        \midrule
        Healthy grass & $57.15 \pm 3.12$ & $61.84 \pm 3.31$ & $\mathbf{69.08 \pm 2.89}$ & $68.25 \pm 4.03$ & $65.30 \pm 3.76$ & $67.75 \pm 4.73$ & $68.00 \pm 3.16$ & $68.63 \pm 2.74$ & $61.74 \pm 4.52$  & $63.00 \pm 4.28$ & $65.16 \pm 5.36$ & $58.88 \pm 5.88$ & $67.97 \pm 4.62$ \\
        Stressed grass & $91.13 \pm 1.38$ & $90.02 \pm 1.16$ & $88.05 \pm 1.65$ & $91.47 \pm 1.62$ & $88.83 \pm 1.76$ & $90.30 \pm 1.83$ & $88.21 \pm 1.93$ & $\mathbf{92.84 \pm 1.94}$ & $92.52 \pm 1.34$  & $88.31 \pm 2.57$ & $91.71 \pm 1.80$ & $89.15 \pm 1.84$ & $92.69 \pm 1.50$ \\
        Artificial turf & $46.81 \pm 13.49$ & $33.85 \pm 4.58$ & $36.98 \pm 17.75$ & $14.01 \pm 9.45$ & $62.64 \pm 33.13$ & $50.35 \pm 33.60$ & $50.47 \pm 13.18$ & $38.66 \pm 21.15$ & $62.86 \pm 16.81$  & $89.75 \pm 5.41$ & $82.19 \pm 11.30$ & $46.09 \pm 15.81$ & $\mathbf{92.86 \pm 4.53}$ \\
        Evergreen trees & $81.81 \pm 4.11$ & $84.39 \pm 3.24$ & $84.92 \pm 7.01$ & $\mathbf{94.27 \pm 0.87}$ & $87.54 \pm 3.53$ & $92.19 \pm 1.37$ & $88.84 \pm 1.93$ & $92.64 \pm 4.57$ & $92.16 \pm 1.97$  & $86.75 \pm 2.09$ & $88.76 \pm 3.72$ & $79.19 \pm 3.35$ & $87.83 \pm 2.02$ \\
        Deciduous trees & $59.38 \pm 2.92$ & $55.82 \pm 3.36$ & $59.42 \pm 1.25$ & $65.82 \pm 4.39$ & $58.69 \pm 2.80$ & $61.83 \pm 4.75$ & $60.60 \pm 5.56$ & $59.60 \pm 3.98$ & $66.48 \pm 5.73$  & $58.56 \pm 6.28$ & $64.32 \pm 4.27$ & $56.99 \pm 4.97$ & $\mathbf{74.79 \pm 2.94}$ \\
        Bare earth & $96.44 \pm 2.37$ & $92.66 \pm 3.02$ & $94.90 \pm 4.48$ & $94.17 \pm 3.67$ & $88.28 \pm 10.83$ & $94.48 \pm 4.60$ & $95.58 \pm 3.36$ & $95.86 \pm 3.94$ & $96.84 \pm 2.58$  & $\mathbf{98.50 \pm 2.48}$ & $96.57 \pm 3.01$ & $94.70 \pm 5.00$ & $98.21 \pm 2.31$ \\
        Water & $24.02 \pm 22.19$ & $1.33 \pm 3.98$ & $1.82 \pm 3.88$ & $0.00 \pm 0.00$ & $0.00 \pm 0.00$ & $2.77 \pm 7.92$ & $32.39 \pm 10.59$ & $12.73 \pm 13.36$ & $29.81 \pm 18.99$  & $44.26 \pm 21.57$ & $51.89 \pm 28.60$ & $7.20 \pm 11.94$ & $\mathbf{64.47 \pm 12.40}$ \\
        Residential buildings & $92.42 \pm 1.42$ & $90.91 \pm 1.12$ & $93.98 \pm 1.18$ & $92.83 \pm 1.02$ & $95.03 \pm 1.66$ & $93.94 \pm 1.02$ & $94.14 \pm 1.13$ & $95.04 \pm 1.77$ & $95.89 \pm 1.15$  & $93.10 \pm 1.39$ & $94.63 \pm 1.47$ & $91.02 \pm 1.59$ & $\mathbf{98.39 \pm 0.38}$ \\
        Non-residential buildings & $\mathbf{99.02 \pm 0.09}$ & $98.59 \pm 0.25$ & $98.38 \pm 0.32$ & $98.39 \pm 0.22$ & $95.72 \pm 3.73$ & $97.61 \pm 0.29$ & $98.42 \pm 0.19$ & $98.57 \pm 0.22$ & $98.89 \pm 0.19$  & $98.36 \pm 0.31$ & $98.46 \pm 0.44$ & $98.43 \pm 0.31$ & $98.98 \pm 0.20$ \\
        Roads & $54.15 \pm 4.59$ & $41.56 \pm 1.94$ & $51.67 \pm 4.79$ & $57.21 \pm 3.35$ & $58.02 \pm 8.92$ & $51.67 \pm 3.89$ & $62.71 \pm 6.77$ & $53.59 \pm 3.44$ & $60.92 \pm 4.06$  & $61.26 \pm 4.66$ & $58.06 \pm 7.50$ & $55.00 \pm 3.00$ & $\mathbf{76.15 \pm 1.98}$ \\
        Sidewalks & $52.90 \pm 3.34$ & $60.20 \pm 1.78$ & $49.81 \pm 6.76$ & $45.01 \pm 3.86$ & $\mathbf{64.43 \pm 3.24}$ & $56.60 \pm 3.09$ & $58.66 \pm 2.47$ & $42.52 \pm 6.09$ & $47.71 \pm 1.60$  & $59.22 \pm 3.78$ & $53.78 \pm 7.16$ & $50.37 \pm 4.45$ & $62.87 \pm 2.44$ \\
        Crosswalks & $0.01 \pm 0.02$ & $0.00 \pm 0.00$ & $0.06 \pm 0.16$ & $0.17 \pm 0.32$ & $0.00 \pm 0.00$ & $\mathbf{4.99 \pm 4.89}$ & $0.80 \pm 1.13$ & $0.00 \pm 0.00$ & $0.27 \pm 0.54$  & $0.88 \pm 1.49$ & $0.50 \pm 1.08$ & $0.01 \pm 0.02$ & $4.11 \pm 1.79$ \\
        Major thoroughfares & $94.85 \pm 1.16$ & $91.26 \pm 0.79$ & $87.52 \pm 2.72$ & $92.10 \pm 1.17$ & $84.39 \pm 7.67$ & $88.46 \pm 2.33$ & $88.38 \pm 1.80$ & $94.43 \pm 1.54$ & $\mathbf{96.37 \pm 0.79}$ & $89.52 \pm 2.56$ & $92.49 \pm 3.06$ & $91.82 \pm 1.97$ & $95.44 \pm 1.32$ \\
        Highways & $93.71 \pm 2.31$ & $82.73 \pm 6.07$ & $78.35 \pm 8.86$ & $\mathbf{95.77 \pm 2.33}$ & $66.60 \pm 17.92$ & $91.25 \pm 4.93$ & $93.07 \pm 4.83$ & $94.01 \pm 6.33$ & $92.48 \pm 4.29$  & $91.81 \pm 4.38$ & $90.23 \pm 6.14$ & $83.56 \pm 7.00$ & $95.69 \pm 2.21$ \\
        Railways & $94.65 \pm 4.47$ & $88.27 \pm 5.08$ & $93.05 \pm 3.50$ & $98.48 \pm 0.68$ & $91.27 \pm 6.14$ & $95.97 \pm 2.35$ & $92.30 \pm 3.23$ & $95.03 \pm 5.99$ & $\mathbf{99.24 \pm 0.71}$ & $93.31 \pm 3.57$ & $96.32 \pm 1.76$ & $94.96 \pm 2.35$ & $97.23 \pm 0.72$ \\
        Paved parking lots & $79.09 \pm 7.65$ & $57.77 \pm 11.45$ & $63.08 \pm 13.83$ & $78.24 \pm 6.04$ & $80.43 \pm 7.43$ & $77.00 \pm 11.01$ & $65.45 \pm 12.73$ & $67.50 \pm 6.12$ & $84.96 \pm 6.92$  & $72.98 \pm 8.60$ & $78.46 \pm 7.78$ & $81.26 \pm 4.96$ & $\mathbf{87.80 \pm 5.21}$ \\
        Unpaved parking lots & $0.00 \pm 0.00$ & $0.00 \pm 0.00$ & $0.00 \pm 0.00$ & $0.00 \pm 0.00$ & $0.00 \pm 0.00$ & $0.00 \pm 0.00$ & $0.41 \pm 1.24$ & $0.00 \pm 0.00$ & $0.00 \pm 0.00$  & $2.07 \pm 3.38$ & $0.00 \pm 0.00$ & $0.00 \pm 0.00$ & $\mathbf{37.45 \pm 27.91}$ \\
        Cars & $78.58 \pm 6.30$ & $50.05 \pm 12.18$ & $73.25 \pm 15.82$ & $47.26 \pm 16.23$ & $56.92 \pm 21.49$ & $69.45 \pm 11.66$ & $76.96 \pm 7.83$ & $55.75 \pm 15.66$ & $\mathbf{87.01 \pm 5.75}$ & $64.33 \pm 10.24$ & $81.52 \pm 9.22$ & $73.83 \pm 9.73$ & $86.01 \pm 4.16$ \\
        Trains & $80.65 \pm 6.75$ & $89.20 \pm 3.51$ & $52.79 \pm 30.65$ & $47.84 \pm 9.74$ & $86.86 \pm 5.47$ & $82.81 \pm 5.02$ & $\mathbf{91.57 \pm 3.91}$ & $71.70 \pm 8.83$ & $43.16 \pm 12.42$  & $89.38 \pm 3.02$ & $87.18 \pm 2.27$ & $49.22 \pm 17.63$ & $91.09 \pm 2.53$ \\
        Stadium seats & $92.09 \pm 3.27$ & $72.00 \pm 9.23$ & $70.04 \pm 11.43$ & $81.82 \pm 8.28$ & $96.72 \pm 1.76$ & $67.72 \pm 11.55$ & $86.25 \pm 5.59$ & $88.24 \pm 6.37$ & $97.85 \pm 2.36$  & $97.76 \pm 1.68$ & $98.29 \pm 1.17$ & $93.17 \pm 4.40$ & $\mathbf{99.74 \pm 0.45}$ \\
        \midrule
        OA (\%) & $87.12 \pm 0.53$ & $84.48 \pm 0.44$ & $84.58 \pm 1.37$ & $86.11 \pm 0.67$ & $85.43 \pm 2.80$ & $86.00 \pm 0.64$ & $87.45 \pm 0.53$ & $86.25 \pm 1.03$ & $88.26 \pm 0.38$  & $87.21 \pm 0.53$ & $87.72 \pm 0.55$ & $85.64 \pm 0.47$ & $\mathbf{91.63 \pm 0.19}$ \\
        AA (\%) & $68.44 \pm 1.97$ & $62.12 \pm 1.40$ & $62.36 \pm 4.10$ & $63.16 \pm 1.77$ & $66.38 \pm 3.88$ & $66.86 \pm 2.65$ & $69.66 \pm 1.43$ & $65.87 \pm 3.46$ & $70.36 \pm 1.55$  & $72.13 \pm 1.56$ & $73.18 \pm 1.80$ & $64.74 \pm 2.10$ & $\mathbf{80.49 \pm 1.73}$ \\
        Kappa & $82.96 \pm 0.72$ & $79.26 \pm 0.65$ & $79.60 \pm 1.86$ & $81.63 \pm 0.93$ & $81.10 \pm 3.45$ & $81.63 \pm 0.85$ & $83.49 \pm 0.72$ & $81.75 \pm 1.36$ & $84.56 \pm 0.51$  & $83.24 \pm 0.69$ & $83.93 \pm 0.73$ & $81.09 \pm 0.64$ & $\mathbf{89.06 \pm 0.24}$ \\
        \bottomrule
    \end{tabular}%
    }
    \endgroup
\end{table*}

\subsection{Feature Visualization and Sensitivity Analysis}
This subsection examines three practical questions beyond aggregate accuracy: whether the learned features are separable, how the model behaves under different training ratios, and how sensitive the method is to the spatial context size. These analyses connect the numerical gains to representation quality, training-ratio sensitivity, and neighborhood scale.

\begin{figure*}[!t]
    \centering
    \includegraphics[width=\textwidth]{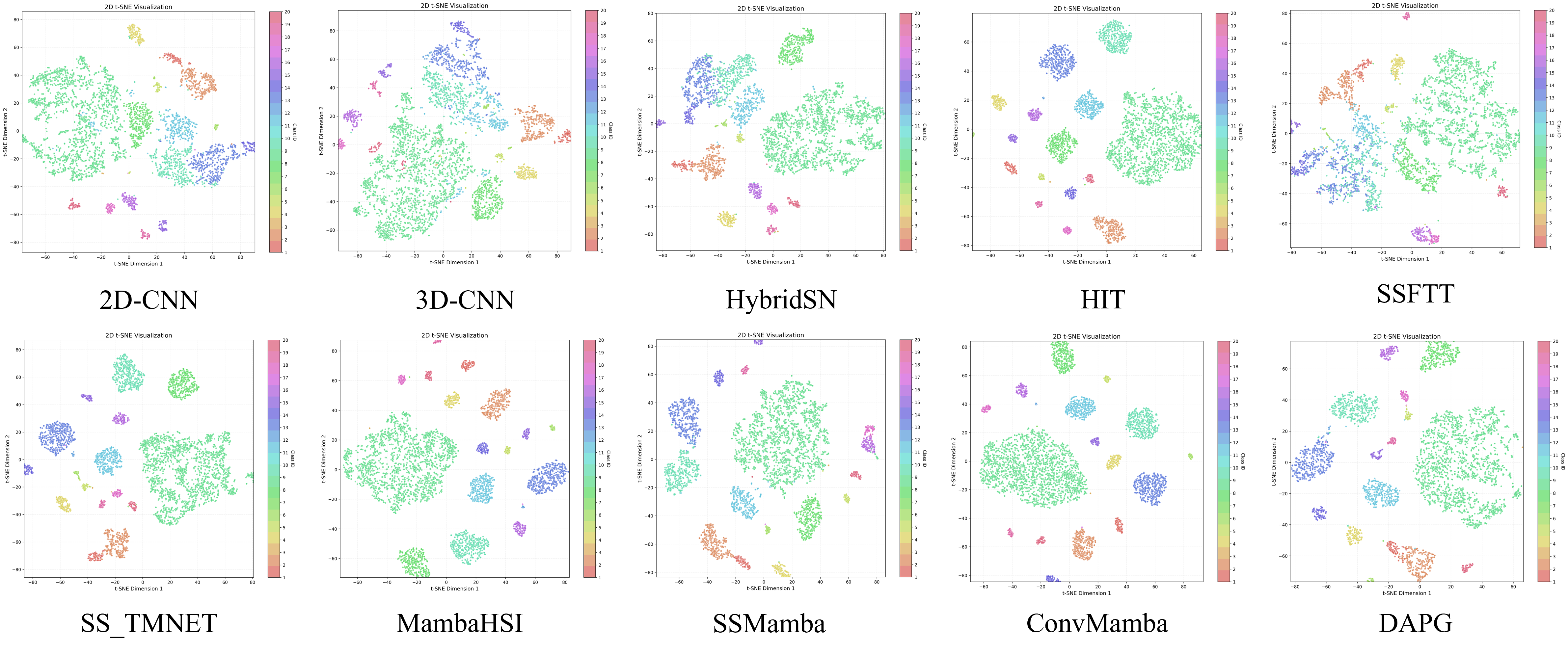}
    \caption{t-SNE visualizations of selected methods on Houston2018. Each panel is labeled with its corresponding method.}\label{fig:houston2018-tsne}
\end{figure*}

\noindent\textit{1) t-SNE visualization:} Fig.~\ref{fig:houston2018-tsne} qualitatively compares the feature distributions produced by the selected methods on Houston2018. The displayed embeddings provide a visual complement to the class-level results in Table~\ref{tab:houston2018-001}. 

\noindent\textit{2) Effect of training ratio on DAPGNet:} Table~\ref{tab:dapgnet-ratio} characterizes the response of DAPGNet to different label budgets. Performance increases as the training ratio grows on all four datasets, while the magnitude of improvement differs across datasets and ratio intervals.

\begin{table*}[!t]
    \centering
    \caption{Effect of training ratio on DAPGNet.}
    \label{tab:dapgnet-ratio}
    \begingroup
    \dapgtableformat
    \begin{adjustbox}{max width=\textwidth}
    \begin{tabularx}{\dapgtablewidth}{llYYYYY}
        \toprule
        Dataset & Metric & $r_1$ & $r_2$ & $r_3$ & $r_4$ & $r_5$ \\
        \midrule
        Indian Pines & OA (\%) & $91.85 \pm 1.18$ & $94.66 \pm 0.51$ & $95.49 \pm 0.41$ & $95.75 \pm 0.18$ & $\mathbf{95.97 \pm 0.22}$ \\
        Indian Pines & AA (\%) & $77.82 \pm 2.95$ & $83.73 \pm 3.79$ & $84.32 \pm 3.45$ & $84.38 \pm 1.47$ & $\mathbf{84.58 \pm 2.08}$ \\
        Indian Pines & Kappa & $90.70 \pm 1.35$ & $93.90 \pm 0.58$ & $94.85 \pm 0.47$ & $95.15 \pm 0.21$ & $\mathbf{95.40 \pm 0.26}$ \\
        \midrule
        WHU-Hi-LongKou & OA (\%) & $91.50 \pm 1.60$ & $95.63 \pm 0.69$ & $97.82 \pm 0.14$ & $98.38 \pm 0.25$ & $\mathbf{98.49 \pm 0.16}$ \\
        WHU-Hi-LongKou & AA (\%) & $67.87 \pm 5.17$ & $82.69 \pm 3.49$ & $91.62 \pm 0.76$ & $93.66 \pm 1.18$ & $\mathbf{94.19 \pm 0.72}$ \\
        WHU-Hi-LongKou & Kappa & $88.70 \pm 2.12$ & $94.21 \pm 0.93$ & $97.13 \pm 0.18$ & $97.86 \pm 0.32$ & $\mathbf{98.02 \pm 0.22}$ \\
        \midrule
        Houston2013 & OA (\%) & $81.74 \pm 1.78$ & $88.29 \pm 1.51$ & $95.13 \pm 0.41$ & $96.92 \pm 0.47$ & $\mathbf{97.80 \pm 0.34}$ \\
        Houston2013 & AA (\%) & $81.35 \pm 2.33$ & $88.60 \pm 1.45$ & $94.93 \pm 0.42$ & $96.69 \pm 0.44$ & $\mathbf{97.49 \pm 0.33}$ \\
        Houston2013 & Kappa & $80.25 \pm 1.93$ & $87.33 \pm 1.63$ & $94.73 \pm 0.44$ & $96.67 \pm 0.51$ & $\mathbf{97.62 \pm 0.37}$ \\
        \midrule
        Houston2018 & OA (\%) & $89.32 \pm 0.36$ & $91.63 \pm 0.19$ & $94.34 \pm 0.14$ & $94.94 \pm 0.28$ & $\mathbf{95.19 \pm 0.28}$ \\
        Houston2018 & AA (\%) & $76.15 \pm 1.82$ & $80.49 \pm 1.73$ & $87.67 \pm 0.48$ & $88.71 \pm 0.96$ & $\mathbf{89.30 \pm 0.66}$ \\
        Houston2018 & Kappa & $86.01 \pm 0.47$ & $89.06 \pm 0.24$ & $92.62 \pm 0.19$ & $93.40 \pm 0.37$ & $\mathbf{93.72 \pm 0.37}$ \\
        \bottomrule
    \end{tabularx}%
    \end{adjustbox}
    \vspace{2pt}
    \begin{minipage}{\textwidth}
    \footnotesize\rmfamily\textit{Note:} For Indian Pines, \(r_1\)--\(r_5\) denote \(5\%, 10\%, 30\%, 50\%, 70\%\); for WHU-Hi-LongKou, they denote \(0.05\%, 0.1\%, 0.3\%, 0.5\%, 0.7\%\); for Houston2013 and Houston2018, they denote \(0.5\%, 1\%, 3\%, 5\%, 7\%\).
    \end{minipage}
    \endgroup
\end{table*}

\noindent\textit{3) Patch-size sensitivity:} Table~\ref{tab:dapgnet-patch} examines the sensitivity of DAPGNet to spatial patch size under the same training protocol. The highest results occur at \(9\times9\) for Indian Pines and WHU-Hi-LongKou, \(13\times13\) for Houston2013, and \(11\times11\) for Houston2018. Larger patches generally yield lower accuracy, with the magnitude of degradation varying across datasets. The analysis characterizes the dataset-dependent influence of spatial context size.

\begin{table*}[!t]
    \centering
    \caption{Patch-size sensitivity.}
    \label{tab:dapgnet-patch}
    \begingroup
    \dapgtableformat
    \begin{adjustbox}{max width=\textwidth}
    \begin{tabularx}{\dapgtablewidth}{llYYYYY}
        \toprule
        Dataset & Metric & $9 \times 9$ & $11 \times 11$ & $13 \times 13$ & $15 \times 15$ & $17 \times 17$ \\
        \midrule
        Indian Pines & OA (\%) & $\mathbf{97.55 \pm 0.37}$ & $97.00 \pm 0.36$ & $95.81 \pm 0.42$ & $94.66 \pm 0.51$ & $92.03 \pm 1.39$ \\
        Indian Pines & AA (\%) & $\mathbf{92.47 \pm 3.21}$ & $89.99 \pm 2.86$ & $87.41 \pm 2.77$ & $83.73 \pm 3.79$ & $76.13 \pm 4.71$ \\
        Indian Pines & Kappa & $\mathbf{97.20 \pm 0.42}$ & $96.58 \pm 0.41$ & $95.22 \pm 0.48$ & $93.90 \pm 0.58$ & $90.90 \pm 1.60$ \\
        \midrule
        WHU-Hi-LongKou & OA (\%) & $\mathbf{96.34 \pm 0.77}$ & $95.94 \pm 0.73$ & $95.76 \pm 0.61$ & $95.63 \pm 0.69$ & $94.78 \pm 0.50$ \\
        WHU-Hi-LongKou & AA (\%) & $\mathbf{85.67 \pm 3.40}$ & $83.94 \pm 3.12$ & $83.46 \pm 2.78$ & $82.69 \pm 3.49$ & $79.01 \pm 2.60$ \\
        WHU-Hi-LongKou & Kappa & $\mathbf{95.16 \pm 1.02}$ & $94.62 \pm 0.98$ & $94.39 \pm 0.81$ & $94.21 \pm 0.93$ & $93.08 \pm 0.67$ \\
        \midrule
        Houston2013 & OA (\%) & $89.11 \pm 1.45$ & $89.09 \pm 0.94$ & $\mathbf{89.36 \pm 0.95}$ & $88.29 \pm 1.51$ & $87.38 \pm 1.70$ \\
        Houston2013 & AA (\%) & $89.23 \pm 1.10$ & $89.36 \pm 0.91$ & $\mathbf{89.55 \pm 0.87}$ & $88.60 \pm 1.45$ & $87.80 \pm 1.77$ \\
        Houston2013 & Kappa & $88.23 \pm 1.57$ & $88.20 \pm 1.01$ & $\mathbf{88.49 \pm 1.03}$ & $87.33 \pm 1.63$ & $86.36 \pm 1.84$ \\
        \midrule
        Houston2018 & OA (\%) & $91.83 \pm 0.17$ & $\mathbf{92.26 \pm 0.13}$ & $92.12 \pm 0.19$ & $91.63 \pm 0.19$ & $90.62 \pm 1.05$ \\
        Houston2018 & AA (\%) & $81.87 \pm 2.20$ & $\mathbf{82.28 \pm 1.75}$ & $82.04 \pm 2.22$ & $80.49 \pm 1.73$ & $77.65 \pm 3.33$ \\
        Houston2018 & Kappa & $89.33 \pm 0.23$ & $\mathbf{89.89 \pm 0.17}$ & $89.71 \pm 0.25$ & $89.06 \pm 0.24$ & $87.74 \pm 1.37$ \\
        \bottomrule
    \end{tabularx}%
    \end{adjustbox}
    \endgroup
\end{table*}

\subsection{Ablation Study}
The ablation study evaluates the contribution of the proposed components and loss formulation. All variants keep the same data splits and training protocol as the corresponding full model, so the observed performance differences quantify the empirical effect associated with each removed module or objective term.

\noindent\textit{1) Module ablation:} Table~\ref{tab:dapgnet-structure} compares the full model with variants that remove the physical-prior branch, replace the adaptive graph with a fixed spatial graph, or remove the physical gate. The full model obtains the highest mean OA, AA, and Kappa on all four datasets. Each removal yields lower mean performance than the full configuration across the evaluated datasets, supporting complementary empirical contributions of the physical-prior branch, adaptive graph construction, and physical gate to the proposed graph learning framework.

\begin{table*}[!t]
    \centering
    \caption{Component ablation.}
    \label{tab:dapgnet-structure}
    \begingroup
    \dapgtableformat
    \begin{adjustbox}{max width=\textwidth}
    \begin{tabularx}{\dapgtablewidth}{llYYY}
        \toprule
        Dataset & Variant & OA (\%) & AA (\%) & Kappa \\
        \midrule
        Indian Pines & Full & $\mathbf{94.66 \pm 0.51}$ & $\mathbf{83.73 \pm 3.79}$ & $\mathbf{93.90 \pm 0.58}$ \\
        Indian Pines & w/o Physical Prior & $94.39 \pm 0.24$ & $82.54 \pm 4.73$ & $93.60 \pm 0.28$ \\
        Indian Pines & w/o Adaptive Graph & $94.13 \pm 0.34$ & $81.92 \pm 3.22$ & $93.29 \pm 0.39$ \\
        Indian Pines & w/o Physical Gate & $94.29 \pm 0.48$ & $81.50 \pm 3.63$ & $93.48 \pm 0.55$ \\
        \midrule
        WHU-Hi-LongKou & Full & $\mathbf{95.63 \pm 0.69}$ & $\mathbf{82.69 \pm 3.49}$ & $\mathbf{94.21 \pm 0.93}$ \\
        WHU-Hi-LongKou & w/o Physical Prior & $95.34 \pm 0.84$ & $82.18 \pm 3.43$ & $93.84 \pm 1.12$ \\
        WHU-Hi-LongKou & w/o Adaptive Graph & $95.22 \pm 0.42$ & $80.52 \pm 1.93$ & $93.66 \pm 0.56$ \\
        WHU-Hi-LongKou & w/o Physical Gate & $95.34 \pm 0.87$ & $81.73 \pm 4.20$ & $93.83 \pm 1.16$ \\
        \midrule
        Houston2013 & Full & $\mathbf{88.29 \pm 1.51}$ & $\mathbf{88.60 \pm 1.45}$ & $\mathbf{87.33 \pm 1.63}$ \\
        Houston2013 & w/o Physical Prior & $87.50 \pm 1.24$ & $87.82 \pm 1.19$ & $86.48 \pm 1.34$ \\
        Houston2013 & w/o Adaptive Graph & $87.30 \pm 1.56$ & $87.62 \pm 1.51$ & $86.26 \pm 1.68$ \\
        Houston2013 & w/o Physical Gate & $87.48 \pm 1.87$ & $87.73 \pm 1.74$ & $86.47 \pm 2.02$ \\
        \midrule
        Houston2018 & Full & $\mathbf{91.63 \pm 0.19}$ & $\mathbf{80.49 \pm 1.73}$ & $\mathbf{89.06 \pm 0.24}$ \\
        Houston2018 & w/o Physical Prior & $91.25 \pm 0.30$ & $78.81 \pm 1.78$ & $88.57 \pm 0.39$ \\
        Houston2018 & w/o Adaptive Graph & $91.41 \pm 0.16$ & $79.43 \pm 1.33$ & $88.77 \pm 0.21$ \\
        Houston2018 & w/o Physical Gate & $91.24 \pm 0.40$ & $79.65 \pm 2.17$ & $88.55 \pm 0.52$ \\
        \bottomrule
    \end{tabularx}%
    \end{adjustbox}
    \endgroup
\end{table*}

\noindent\textit{2) Loss ablation:} Table~\ref{tab:encoder-loss} compares cross-entropy alone, cross-entropy with auxiliary supervision, and the complete objective with spectral smoothness regularization. Auxiliary supervision produces different mean effects across datasets. The complete objective obtains the highest mean OA, AA, and Kappa on every dataset, indicating that the combined loss formulation provides the strongest empirical results among the evaluated objectives.

\begin{table*}[!t]
    \centering
    \caption{Loss ablation.}
    \label{tab:encoder-loss}
    \begingroup
    \dapgtableformat
    \begin{adjustbox}{max width=\textwidth}
    \begin{tabularx}{\dapgtablewidth}{llYYY}
        \toprule
        Dataset & Metric & CE & CE + AuxCE & CE + AuxCE + Phy \\
        \midrule
        Indian Pines & OA (\%) & $94.41 \pm 0.55$ & $94.29 \pm 0.57$ & $\mathbf{94.66 \pm 0.51}$ \\
        Indian Pines & AA (\%) & $82.78 \pm 3.81$ & $82.37 \pm 3.22$ & $\mathbf{83.73 \pm 3.79}$ \\
        Indian Pines & Kappa & $93.62 \pm 0.62$ & $93.48 \pm 0.65$ & $\mathbf{93.90 \pm 0.58}$ \\
        \midrule
        WHU-Hi-LongKou & OA (\%) & $94.74 \pm 0.93$ & $95.37 \pm 0.76$ & $\mathbf{95.63 \pm 0.69}$ \\
        WHU-Hi-LongKou & AA (\%) & $79.08 \pm 5.14$ & $82.01 \pm 3.38$ & $\mathbf{82.69 \pm 3.49}$ \\
        WHU-Hi-LongKou & Kappa & $93.02 \pm 1.25$ & $93.87 \pm 1.02$ & $\mathbf{94.21 \pm 0.93}$ \\
        \midrule
        Houston2013 & OA (\%) & $87.22 \pm 1.57$ & $87.89 \pm 1.80$ & $\mathbf{88.29 \pm 1.51}$ \\
        Houston2013 & AA (\%) & $87.61 \pm 1.64$ & $88.23 \pm 1.67$ & $\mathbf{88.60 \pm 1.45}$ \\
        Houston2013 & Kappa & $86.19 \pm 1.69$ & $86.91 \pm 1.94$ & $\mathbf{87.33 \pm 1.63}$ \\
        \midrule
        Houston2018 & OA (\%) & $91.47 \pm 0.30$ & $91.49 \pm 0.36$ & $\mathbf{91.63 \pm 0.19}$ \\
        Houston2018 & AA (\%) & $80.30 \pm 2.00$ & $79.77 \pm 1.81$ & $\mathbf{80.49 \pm 1.73}$ \\
        Houston2018 & Kappa & $88.85 \pm 0.39$ & $88.88 \pm 0.48$ & $\mathbf{89.06 \pm 0.24}$ \\
        \bottomrule
    \end{tabularx}%
    \end{adjustbox}
    \endgroup
\end{table*}
\subsection{Computational Complexity and Efficiency}
Table~\ref{tab:houston2018-model-efficiency} reports the parameter count, FLOPs, whole-test inference time, and training time of representative models on Houston2018 with a \(15\times15\) patch.

\begin{table*}[!t]
    \centering
    \caption{Computational complexity and timing of representative models on Houston2018 with a \(15\times15\) patch.}
    \label{tab:houston2018-model-efficiency}
    \begingroup
    \dapgtableformat
    \setlength{\tabcolsep}{2.2pt}
    \renewcommand{\arraystretch}{1.12}
    \begin{adjustbox}{max width=\textwidth}
    \begin{tabular}{l*{13}{c}}
        \toprule
        Metric & 2D-CNN & 3D-CNN & HybridSN & SSFTT & HIT & SS-TMNet & MambaHSI & ConvMamba & SSMamba & S2VNet & DAPGNet & GraphMamba & DBMGNet \\
        \midrule
        Params (M) & \(1.7605\) & \(1.1413\) & \(1.5224\) & \(0.1267\) & \(12.5808\) & \(59.3294\) & \(0.3697\) & \(0.4560\) & \(18.4138\) & \(0.1041\) & \(1.0520\) & \(0.9846\) & \(0.3802\) \\
        FLOPs (G) & \(0.0378\) & \(0.3779\) & \(0.1037\) & \(0.0159\) & \(0.6447\) & \(0.9421\) & \(0.0500\) & \(0.1238\) & \(0.8549\) & \(0.0094\) & \(0.8948\) & \(0.0459\) & \(0.0039\) \\
        Inference Time (s) & \(27.24\) & \(78.53\) & \(25.98\) & \(28.21\) & \(199.22\) & \(463.26\) & \(282.25\) & \(77.48\) & \(701.25\) & \(63.40\) & \(373.65\) & \(699.87\) & \(1683.56\) \\
        Training Time (s) & \(118.21\) & \(207.48\) & \(162.21\) & \(192.82\) & \(1166.45\) & \(2026.99\) & \(894.74\) & \(368.97\) & \(1655.55\) & \(432.59\) & \(875.26\) & \(1381.26\) & \(1211.15\) \\
        \bottomrule
    \end{tabular}
    \end{adjustbox}
    \endgroup
\end{table*}

The parameter counts show that DAPGNet uses a moderate model size. With 1.0520M parameters, it is smaller than the convolutional baselines 2D-CNN, 3D-CNN, and HybridSN, and is much smaller than HIT, SS-TMNet, and SSMamba. Its parameter count is also close to GraphMamba, indicating that the proposed relation-level physical-prior mechanism does not rely on an oversized network. Although several lightweight baselines, including SSFTT, MambaHSI, ConvMamba, S2VNet, and DBMGNet, contain fewer parameters, their Houston2018 accuracy remains lower than DAPGNet under the same experimental setting. DAPGNet therefore achieves its accuracy gain under a moderate parameter budget.

The timing results further reveal the distinction between model size and operational cost. DAPGNet requires higher FLOPs and longer inference time than shallow convolutional models and some lightweight sequence models, mainly because the adaptive graph constructor evaluates candidate relations and the graph diffusion layers perform edge-biased attention with physical-prior gating. At the same time, its training time is lower than HIT, SS-TMNet, SSMamba, GraphMamba, and DBMGNet, and its inference time is lower than SSMamba, GraphMamba, and DBMGNet. These results indicate that the relation-level physical-prior design introduces additional computation while keeping the overall cost within a practical range for patch-based HSI classification.

\section{Discussion}
The experimental results collectively indicate that extending the physical prior from node representation to relation-level graph learning provides consistent benefits for hyperspectral image classification. Across the four benchmark datasets, DAPGNet achieves favorable OA, AA, and Kappa values relative to the representative comparison methods. The more pronounced gains in AA suggest that the proposed design is also beneficial to class-balanced recognition. Performance degradation is observed after removing the multiscale spectral physical-prior encoder, adaptive graph construction, or physical gate, indicating that these components make complementary contributions to relation-level graph learning. Collectively, these findings support the central premise of DAPGNet: structural regularities contained in contiguous spectra can be more fully utilized when they jointly participate in node representation, graph topology estimation, and layer-wise graph diffusion.

From a graph-topology perspective, feature affinity can be ambiguous around mixed pixels and heterogeneous regions. DAPGNet jointly considers shallow spectral--spatial features, physical-prior consistency, and spatial distance during candidate selection and edge-weight estimation. This design provides an additional relation cue and reduces the dependence of topology estimation on a single feature space. The ablation results support the contribution of the physical-prior-aware topology. However, graph homophily and edge purity have not yet been directly quantified. Accordingly, the present evidence supports the effectiveness of the proposed mechanism but does not quantify the reduction in erroneous connections. After topology construction, the learned edge weights remain active as additive biases in sparse attention, thereby linking neighbor selection with message weighting. The physical gate further balances graph-aggregated features and projected physical-prior features at each diffusion layer. The physical prior consequently remains involved in topology estimation, attention regulation, and node-state updating. This continuous path is consistent with the complementary contributions of adaptive graph construction and physical gating observed in the ablation study.

The present formulation follows a structure-constrained route to physics-guided learning. It derives the physical prior mainly from the ordering, local continuity, spectral smoothness, and multiscale variation of contiguous spectral measurements. Center wavelengths and sensor spectral response functions are not explicitly incorporated. Consequently, nonuniform spectral sampling and sensor-dependent observation processes are not directly modeled. The three spectral kernel widths describe ranges in band indices and do not correspond to fixed physical wavelength intervals. Future work may extend DAPGNet with wavelength-aware spectral operators and sensor-response-aware prior encoding. When reliable task-specific knowledge is available, equation- or model-driven constraints derived from radiative transfer, spectral mixing, or endmember--abundance relationships may also be incorporated into relation-level graph learning. Since HSI classification lacks a universal governing equation across sensors, scenes, and applications, such physical constraints should be selected according to the sensing configuration and target task. Moreover, graph construction remains confined to individual local patches, so cross-patch long-range relations are not directly modeled. Although the two-stage graph construction strategy restricts computationally expensive learnable edge scoring to a reduced candidate set, coarse pairwise relation estimation still scales approximately quadratically with the number of nodes. Local--global or hierarchical graphs, approximate neighbor search, and optimized sparse operators are promising directions for extending the framework to larger spatial contexts.
\section{Conclusion}

This work demonstrates that structure-constrained spectral physical priors contribute to HSI classification beyond representation learning by regulating graph topology and layer-wise propagation. Based on this view, DAPGNet couples multiscale contiguous-spectral prior encoding with physical-prior-aware graph construction and edge-biased physics-gated diffusion. The resulting graph learning process uses the prior to guide neighbor selection, attention allocation, and node updates while retaining data-driven spectral--spatial features. Experiments on four HSI datasets show that DAPGNet obtains the best OA, AA, and Kappa among representative CNN-, Transformer-, state-space/Mamba-, and graph-based baselines, with AA gains of 3.64--7.31 percentage points over the strongest competing method. Component ablations show that the physical-prior branch, adaptive graph constructor, and physical gate provide complementary support for relation-level graph learning, while the loss ablation supports auxiliary supervision and spectral smoothness regularization as useful training signals. Sensitivity analyses further characterize the effects of training ratio and patch size under different supervision and spatial-context settings. The current formulation remains bounded by its structure-constrained prior: sensor response, atmospheric transfer, and material-specific radiative processes are not explicitly modeled, and the patch-level graph restricts global scene relations. Future work will incorporate richer physical metadata and evaluate graph construction under larger-scale inference settings.

\bibliographystyle{IEEEtran}
\bibliography{references}

\end{document}